%% file: main.tex
\pdfoutput=1

\documentclass[11pt]{article}

\usepackage[final]{acl}

\usepackage{times}
\usepackage{latexsym}

\usepackage[T1]{fontenc}

\usepackage[utf8]{inputenc}

\usepackage{microtype}

\usepackage{inconsolata}

\usepackage{graphicx}

\usepackage{mdframed}
\usepackage{booktabs}
\usepackage{multirow}
\usepackage{subcaption}
\usepackage{verbatim}
\usepackage{amssymb}
\usepackage{pifont}
\usepackage{tablefootnote}
\usepackage{enumitem}
\usepackage{amsmath}
\usepackage{amssymb}
\usepackage{amsfonts}
\usepackage{dsfont}
\usepackage{bm}

\newcommand*{\system}{\textcolor[HTML]{E8000B}}
\newcommand*{\example}{\textcolor[HTML]{3CB44B}}
\newcommand*{\question}{\textcolor[HTML]{911EB4}}

\newcommand{\cmark}{\ding{51}}%
\newcommand{\xmark}{\ding{55}}%
\newcommand{\tri}{\ding{115}}%
\newcommand*{\greencheckmark}{\color[HTML]{3CB44B}{\cmark}}
\newcommand*{\redx}{\color[HTML]{E8000B}{\xmark}}
\newcommand*{\orangetriangle}{\color{orange}{\tri}}

\title{Medical Adaptation of Large Language and Vision-Language Models:\\Are We Making Progress?}

\author{
  \textbf{Daniel P. Jeong\textsuperscript{1}},
  \textbf{Saurabh Garg\textsuperscript{1,2}},
  \textbf{Zachary C. Lipton\textsuperscript{1,4}},
  \textbf{Michael Oberst\textsuperscript{3,4}}
\\
  \textsuperscript{1}Machine Learning Department, Carnegie Mellon University\\
  \textsuperscript{2}Mistral AI\\
  \textsuperscript{3}Department of Computer Science, Johns Hopkins University\\
  \textsuperscript{4}Abridge AI\\
  \texttt{\{danielje,sgarg2,zlipton\}@cs.cmu.edu,moberst@jhu.edu}
\\
  \small{
    \textbf{Correspondence:} \href{mailto:danielje@cs.cmu.edu}{danielje@cs.cmu.edu}
  }
}

\begin{document}
\maketitle
\begin{abstract}
\input{latex/sections/0_abstract}
\end{abstract}

\section{Introduction}\label{sec:intro}
\input{latex/sections/1_intro}

\section{Related Work}\label{sec:related-work}
\input{latex/sections/2_related-work}

\section{Experimental Setup}\label{sec:eval-setup}
\input{latex/sections/3_eval-setup}

\section{Results}\label{sec:results}
\input{latex/sections/4_results}

\section{Discussion and Conclusion}\label{sec:discussion}
\input{latex/sections/5_discussion}

\section{Limitations}\label{sec:limitations}
\input{latex/sections/6_limitations}

\section*{Acknowledgments}
We gratefully acknowledge DARPA (FA8750-23-2-1015), ONR (N00014-23-1-2368), NSF (IIS2211955), UPMC, Highmark Health, Abridge, Ford Research, Mozilla, the PwC Center, Amazon AI, JP Morgan Chase, the Block Center, the Center for Machine Learning and Health, and the CMU Software Engineering Institute (SEI) via Department of Defense contract FA8702-15-D-0002, for their generous support of our research.

\clearpage
\bibliography{ref}
\clearpage

\appendix

\renewcommand\thetable{A\arabic{table}}
\renewcommand\thefigure{A\arabic{figure}}
\setcounter{table}{0}
\setcounter{figure}{0}
\section{Additional Details on Datasets}\label{sec:details-datasets}
\input{latex/sections/appendix_datasets}

\renewcommand\thetable{B\arabic{table}}
\renewcommand\thefigure{B\arabic{figure}}
\setcounter{table}{0}
\setcounter{figure}{0}
\section{Additional Details on Model-Specific Prompt Selection}\label{sec:details-prompting-selection}
\input{latex/sections/appendix_prompt-selection}

\renewcommand\thetable{C\arabic{table}}
\renewcommand\thefigure{C\arabic{figure}}
\setcounter{table}{0}
\setcounter{figure}{0}
\section{Additional Details on Zero-/Few-shot Prompting}\label{sec:details-prompting}
\input{latex/sections/appendix_prompting}

\renewcommand\thetable{D\arabic{table}}
\renewcommand\thefigure{D\arabic{figure}}
\setcounter{table}{0}
\setcounter{figure}{0}
\section{Additional Results for the Zero-/Few-Shot Prompting Evaluations with Greedy Decoding}\label{sec:prompting-results-appendix}
\input{latex/sections/appendix_results}

\renewcommand\thetable{E\arabic{table}}
\renewcommand\thefigure{E\arabic{figure}}
\setcounter{table}{0}
\setcounter{figure}{0}
\section{Results for the Zero-/Few-Shot Prompting Evaluations with Constrained Decoding}\label{sec:constrained-decoding}
\input{latex/sections/appendix_constrained_results}

\end{document}

%% file: latex/sections/0_abstract.tex
Several recent works seek to develop foundation models specifically for medical applications, 
adapting general-purpose
large language models (LLMs) 
and vision-language models (VLMs) 
via continued pretraining
on publicly available biomedical corpora. 
These works typically claim that such domain-adaptive pretraining (DAPT) 
improves performance on downstream medical tasks,
such as answering medical licensing exam questions.  
In this paper, we compare seven public ``medical'' LLMs and two VLMs
against their corresponding base models,
arriving at a different conclusion: 
all medical VLMs and nearly all medical LLMs 
fail to consistently improve over their base models 
in the zero-/few-shot prompting regime 
for medical question-answering (QA) tasks.
For instance, across the tasks and model pairs we consider in the 3-shot setting, 
medical LLMs only outperform their base models in 12.1\% of cases,
reach a (statistical) tie in 49.8\% of cases, 
and are significantly worse than their base models 
in the remaining 38.2\% of cases.
Our conclusions are based on  
(i) comparing each medical model head-to-head, directly 
against the corresponding base model;
(ii) optimizing the prompts for each model separately;
and (iii) accounting for statistical uncertainty in comparisons. 
While these basic practices are not consistently adopted in the literature,
our ablations show that they substantially impact conclusions.
Our findings suggest that state-of-the-art general-domain models 
may already exhibit strong medical knowledge and reasoning capabilities, 
and offer recommendations to strengthen the conclusions of future studies.

\textit{This version was published at EMNLP 2024. In the \href{https://arxiv.org/abs/2411.08870}{extended version} of our paper, we also include the results on closed-ended QA tasks based on clinical notes in addition to medical-exam-style QA, as well as a comparison of performance when using medical versus general-domain models as an initialization for downstream supervised fine-tuning.}

%% file: latex/sections/1_intro.tex
\begin{figure*}[t!]
    \centering
    \begin{tabular}{c@{}c@{\hskip 3pt}c@{}c}
        \begin{subfigure}{0.03\linewidth}
            \makebox[\linewidth]{\raisebox{55pt}{{(a)}}}
        \end{subfigure} &
        \begin{subfigure}[c]{0.5\linewidth}
            \includegraphics[width=\linewidth]{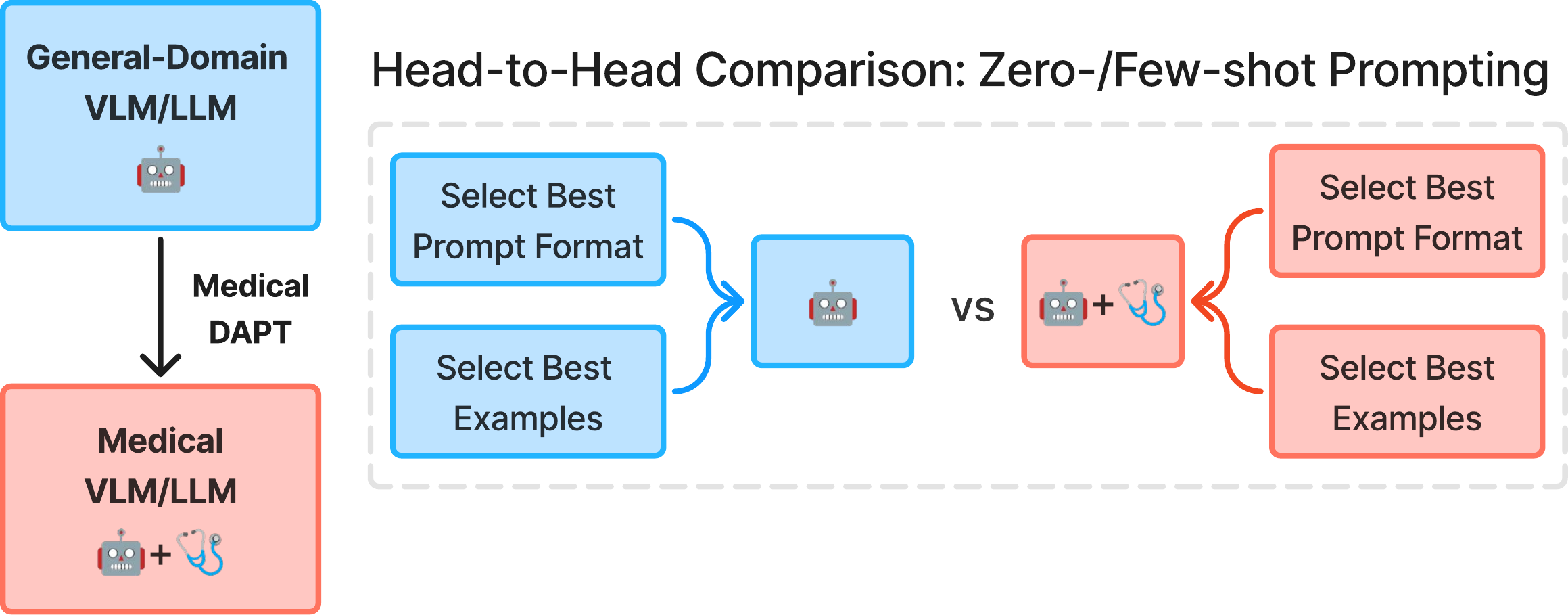}
        \end{subfigure} &
        \begin{subfigure}{0.03\linewidth}
            \makebox[\linewidth]{\raisebox{55pt}{{(b)}}}
        \end{subfigure} &
        \begin{subfigure}[c]{0.42\linewidth}
            \includegraphics[width=\linewidth]{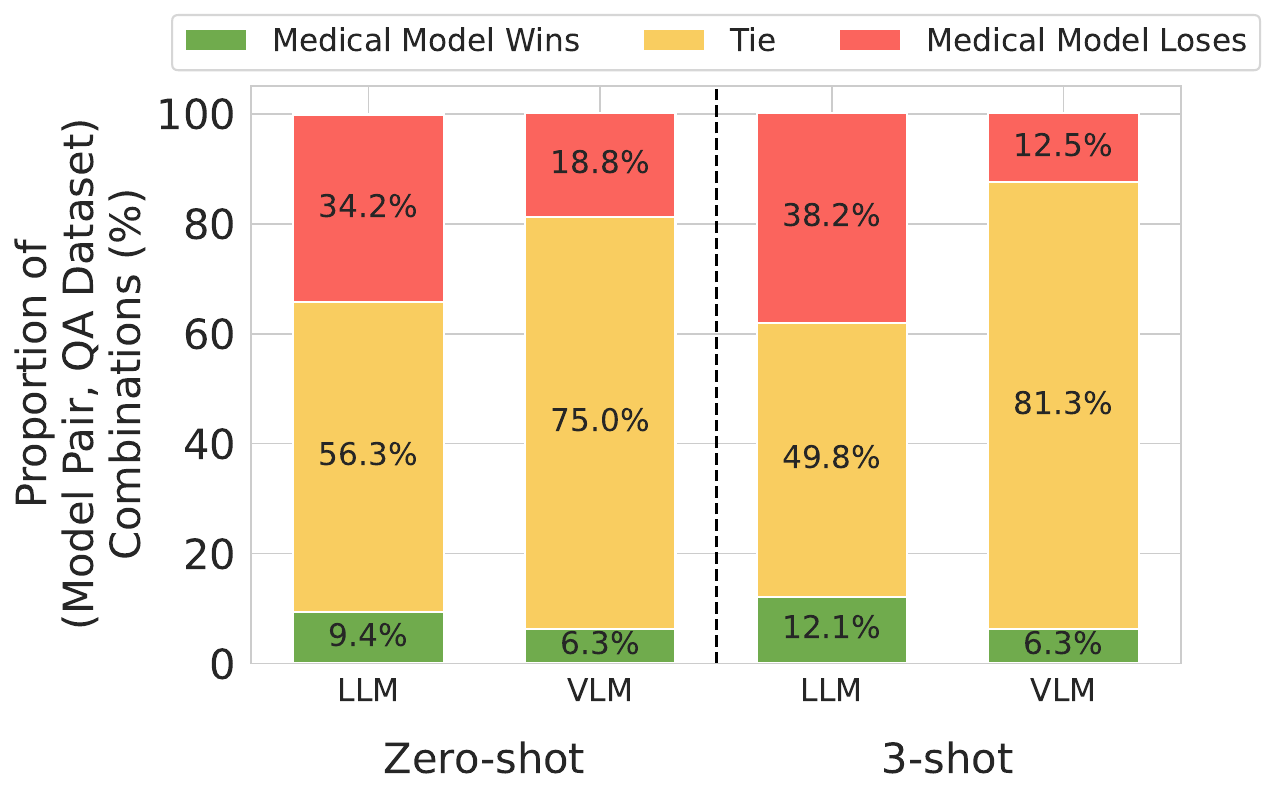}
        \end{subfigure}
    \end{tabular}
    \vspace{-5pt}
    \caption{Medical LLMs and VLMs trained via domain-adaptive pretraining (DAPT) show limited improvement over their general-domain counterparts. (a) Overview of our head-to-head evaluation approach for each pair of general-domain (\textcolor{cyan}{blue}) and medically adapted LLM/VLM (\textcolor{red}{red}). 
    (b) Win/tie/loss rate (\%) of medical models vs. their corresponding base models across all (model pair, QA dataset) combinations. 
    Win rate refers to the proportion of (model pair, QA dataset) combinations where a medical model shows a statistically significant improvement.}
    \label{fig:intro-fig}
\end{figure*}

Recent advances in autoregressive large language models (LLMs)
and vision-language models (VLMs) have attracted interest
from practitioners in medicine, 
where these models hold great potential to transform
various aspects of clinical practice 
(e.g., medical diagnosis, information retrieval from clinical documents, patient triaging) 
\citep{how-ai-can-transform,generalist-medai}.
State-of-the-art performance on various medical benchmarks 
is typically achieved by massive-scale closed-source models,
such as \textsc{GPT-4} \citep{gpt4,gpt4v}, \textsc{Med-Gemini} \citep{med-gemini,med-gemini-2}, 
and \textsc{Med-PaLM} \citep{llm-clinical,med-palm-2,med-palm-m}, 
often performing on par with humans on medical licensing exams 
and open-ended consumer health question-answering (QA) tasks.
However, the general lack of transparency in these models, 
high API usage costs, and patient data privacy concerns 
make their integration into routine clinical workflows challenging \citep{llm-hipaa}. 

To address such concerns, recent works 
have proposed cheaper, open-source alternatives
through \textit{domain-adaptive pretraining}
\citep[DAPT;][]{dapt}, where a pretrained open-source general-domain model---such 
as \textsc{Llama} \citep{llama-1,llama-2,llama-3} or \textsc{Mistral} \citep{mistral} in the language space, 
and \textsc{LLaVA} \citep{llava} or \textsc{Open-Flamingo} \citep{open-flamingo} in the vision-language space---is continually pretrained on biomedical (image-)text corpora 
from public sources such as PubMed and medical textbooks. 
While some prior works show that medical models pretrained from scratch 
only using domain-specific corpora can outperform those trained via DAPT, 
both in the context of BERT-style encoder-only models \citep{bert,pubmedbert,gatortron} 
and decoder models \citep{galactica,biogpt,do-we-still-need-clinical-lms,biomedlm}, 
the DAPT approach has become common practice, 
resulting in a trend where the release of a more capable general-domain model
is typically followed by the release of its medical counterpart. 

Despite the widespread adoption of medical DAPT, 
the claimed improvements in performance are worth scrutinizing. 
While the story is intuitive, more recent base models 
(e.g., \textsc{Llama-3-8B} \citep{llama-3}) 
already exhibit strong off-the-shelf performance 
on medical benchmarks without any adaptation (e.g., Open Medical LLM Leaderboard \citep{open-medical-llm-leaderboard}), 
and given a lack of transparency about the pretraining corpora 
used to train the general-domain model in the first place, 
they may already be trained on relevant medical text.

Perhaps more concerning is the lack of apples-to-apples comparisons in the literature. 
First, medical models resulting from DAPT are often only compared 
against baselines with different architectures 
(e.g., \textsc{Clinical-Camel-70B} \citep{clinical-camel} vs. \textsc{GPT-4} \citep{gpt4}) 
and under inconsistent evaluation setups 
(e.g., \textsc{MediTron-70B} \citep{meditron} fine-tuned on MedQA \citep{medqa} 
vs. non-fine-tuned \textsc{Med42-v1-70B} \citep{med42-v1}), 
which can confound the interpretation of results. 
Second, the common practice of using a single, 
fixed prompting setup (e.g., prompt format, choice of few-shot examples) 
for all models under evaluation also warrants concern,
as LLM/VLM behavior is extremely sensitive to such design decisions 
\citep{how-can-we-know-what-lms-know,calibrate-before-use,open-clinical-llms-sensitive}, 
and the ``optimal'' choice of such details rarely correlates 
between different models \citep{quantifying-lm-prompt-design}.

In this paper, we perform an apples-to-apples comparison that addresses these concerns, 
comparing seven medical LLMs and two medical VLMs against their general-domain base models. 
We find that, for all but one LLM pair---\textsc{BioMistral-7B} \citep{biomistral} vs. \textsc{Mistral-7B-Instruct-v0.1} \citep{mistral}, 
a pair of models that performs fairly poorly in absolute terms---the open-source medical LLMs and VLMs that we evaluate do not consistently improve over their general-domain counterparts on various medical (visual) QA tasks (Figure \ref{fig:intro-fig}). 
We compare several pairs of general-domain and medically adapted LLMs/VLMs (see Table \ref{tab:models}),
\textbf{whose only differences lie in medical DAPT} (i.e., one model is the base model, from which the other is derived via medical DAPT). For each pair, we compare their performances from zero-/few-shot prompting \citep{gpt-2,gpt-3}, after independently selecting the ``best'' prompt format and few-shot examples for each model based on the validation set and accounting for statistical uncertainty in model comparison. 

\begin{table*}[t!]
    \centering
    \caption{Summary of open-source autoregressive VLM and LLM pairs used for evaluation.}
    \label{tab:models}
    \resizebox{\linewidth}{!}{
    \begin{tabular}{@{}c@{\hskip 7pt}l@{\hskip 7pt}l@{\hskip 7pt}l@{\hskip 3pt}}
        \toprule
        Model Class & General Domain & Medical Domain & Medical Adaptation Corpora \\
        \midrule
        \multirow{12}{*}{LLM} & \textsc{Llama-3-70B-Instruct}~\citep{llama-3} & \textsc{OpenBioLLM-70B}~\citep{OpenBioLLMs} & Undisclosed \\
        \cmidrule(){2-4}
        & \multirow{2}{*}{\textsc{Llama-2-70B}~\citep{llama-2}} & \multirow{2}{*}{\textsc{MediTron-70B}~\citep{meditron}} & Clinical Practice Guidelines (e.g., CDC, WHO) \\
        & & & PubMed Articles~\citep[S2ORC;][]{s2orc} \\
        \cmidrule(){2-4}
        & \multirow{3}{*}{\textsc{Llama-2-70B}~\citep{llama-2}} & \multirow{3}{*}{\textsc{Clinical-Camel-70B}~\citep{clinical-camel}} & ShareGPT \\
        & & & 20k PubMed Articles Published Before 2021 \\
        & & & Random 4k Subset of MedQA~\citep{medqa} \\
        \cmidrule(){2-4}
        & \textsc{Llama-3-8B} \citep{llama-3} & \textsc{OpenBioLLM-8B} \citep{OpenBioLLMs} & Undisclosed \\
        \cmidrule(){2-4}
        & \multirow{2}{*}{\textsc{Llama-2-7B} \citep{llama-2}} & \multirow{2}{*}{\textsc{MediTron-7B} \citep{meditron}} & Clinical Practice Guidelines (e.g., CDC, WHO) \\
        & & & PubMed Articles \citep[S2ORC;][]{s2orc} \\
        \cmidrule(){2-4}
        & \textsc{Mistral-7B-Instruct-v0.1} \citep{mistral} & \textsc{BioMistral-7B} \citep{biomistral} & PubMed Articles (PMC Open Access Subset) \\
        \cmidrule(){2-4}
        & \textsc{Llama-2-7B-Chat} \citep{llama-2} & \textsc{BioMedGPT-LM-7B} \citep{biomedgpt} & PubMed Articles \citep[S2ORC;][]{s2orc} \\
        \midrule
        \multirow{3}{*}{VLM} & \textsc{LLaVA-v0-7B}~\citep{llava} & \textsc{LLaVA-Med-7B}~\citep{llava-med} & PubMed Articles~\citep[PMC-15M;][]{biomedclip} \\
        \cmidrule(){2-4}
        & \multirow{2}{*}{\textsc{Open-Flamingo-9B}~\citep{open-flamingo}} & \multirow{2}{*}{\textsc{Med-Flamingo-9B}~\citep{med-flamingo}} & Medical Textbooks~\citep[MTB;][]{med-flamingo} \\
        & & & PubMed Articles \citep[PMC-OA;][]{pmcclip}\\
        \bottomrule
    \end{tabular}
    }
\end{table*}

Our findings (Section \ref{sec:results}) suggest that state-of-the-art general-domain models 
may already exhibit strong medical knowledge and reasoning capabilities
that can be leveraged effectively when prompted appropriately. 

Our main contributions can be summarized as follows:
\begin{enumerate}[topsep=0.5ex,itemsep=-0.5ex]
    \item We provide a comprehensive head-to-head comparison between state-of-the-art general-domain LLMs/VLMs and their medical DAPT counterparts on various medical (visual) QA benchmarks, to investigate the effectiveness of DAPT for medical specialization.
    \item We find that after optimizing the prompts for medical and general-domain models independently, all medical VLMs and nearly all medical LLMs that we evaluate fail to consistently improve over their corresponding general-domain base models.
    \item We show that using a single, fixed prompt format and choice of few-shot examples for all models without testing for statistical significance can lead to overly optimistic conclusions about the benefits from medical DAPT.
\end{enumerate}

%% file: latex/sections/2_related-work.tex
DAPT \citep{dapt} is a transfer learning approach, 
where a pretrained model is further pretrained on domain-specific data
for better alignment to a target domain of interest (e.g., medicine, law). 
Several studies show that language models trained via DAPT 
often outperform their general-domain counterparts on domain-specific tasks,
such as claim detection from blog posts \citep{chakrabarty-etal-2019-imho}, 
named entity recognition from German novels \citep{german-novel}, 
and judgment prediction for legal cases \citep{lawformer}.
In the medical domain, prior works based on BERT-style encoder-only language models \citep{bert}, 
such as \textsc{BioBERT} \citep{biobert} and \textsc{ClinicalBERT} \citep{clinical-bert-alsentzer}, 
show that medical DAPT improves fine-tuning performance on tasks 
such as medical concept extraction from patient reports \citep{i2b2-2010}, 
identification of gene-disease relations from PubMed abstracts \citep{ncbi-disease,gad,chemprot},
and natural language inference on clinical notes \citep{mednli}. 
 
More recent works suggest that decoder-based autoregressive LLMs and VLMs trained via medical DAPT 
also show strong performance on various medical tasks.
Medical LLMs such as \textsc{MediTron} \citep{meditron}, adapted from \textsc{Llama-2} \citep{llama-2};
and \textsc{BioMistral} \citep{biomistral}, adapted from \textsc{Mistral-7B-Instruct-v0.1} \citep{mistral}; 
perform well on knowledge-intensive QA tasks based on medical licensing 
and academic exams \citep{medqa,medmcqa,mmlu} and PubMed abstracts \citep{pubmedqa}. 
Medical VLMs such as \textsc{LLaVA-Med} \citep{llava-med}, 
adapted from \textsc{LLaVA} \citep{llava}; 
and \textsc{Med-Flamingo} \citep{med-flamingo}, adapted from \textsc{Open-Flamingo} \citep{open-flamingo}; also perform well on visual QA tasks based on radiology \citep{vqa-rad,slake} 
and pathology images \citep{pvqa} 
and academic exams \citep{mmmu}.
These encouraging results have established DAPT as a go-to approach 
for training a medically specialized model,
a conclusion that we re-examine in this work.

%% file: latex/sections/3_eval-setup.tex
To investigate the effectiveness of medical DAPT in improving zero-/few-shot performance, 
we compare 7 medical LLMs and 2 medical VLMs against their general-domain counterparts in \textit{pairs} (Figure \ref{fig:intro-fig}(a)), on 13 textual QA datasets and 8 visual QA datasets, respectively. 
The models in each pair are \textit{exactly identical} in model architecture and scale, and their only difference lies in whether they were additionally pretrained on medical data. We also note that while some of datasets used for evaluation contain both closed-ended (i.e., has clear ground-truth answers) and open-ended questions, we focus our evaluations on the former, where an objective, quantitative assessment of medical knowledge and reasoning capabilities is possible. For reproducibility of our results, we open-source the source code used for all of our evaluations described below via our GitHub repository\footnote{\href{https://github.com/taekb/eval-medical-dapt}{https://github.com/taekb/eval-medical-dapt}}.

\paragraph{Models.} In Table \ref{tab:models}, we provide a summary of all of the LLM and VLM pairs that we use for evaluation, along with details about the pretraining corpora used for adaptation to the medical domain. 
For \textsc{LLaVA} \citep{llava}, we use the very first version (v0) that uses \textsc{Vicuna-v0} \citep{vicuna} as the LLM backbone, as \textsc{LLaVA-Med} \citep{llava-med} was adapted from that particular version.
For all models, we use the checkpoints made available via HuggingFace. 
In all experiments, we generate predictions from each model via (i) greedy decoding (i.e., sampling with temperature $T=0$) and (ii) constrained decoding. For constrained decoding, we constrain the token vocabulary to be one of the answer choice letters (e.g., one of [``A'', ``B'', ``C'', ``D''] for a four-choice QA dataset) and treat the answer choice with the highest token probability as a given model's prediction.

\paragraph{Textual QA Datasets.} For textual QA, we use MedQA \citep{medqa}, MedMCQA \citep{medmcqa}, PubMedQA \citep{pubmedqa}, and MMLU-Medical \citep{mmlu} for evaluation. 
MMLU-Medical refers to a subset of MMLU corresponding to 9 subjects related to medicine: anatomy, clinical knowledge, college biology, college medicine, high school biology, medical genetics, nutrition, professional medicine, and virology. 
For MedQA, we use the official train-validation-test splits as provided through BigBio \citep{bigbio}. We note that MedQA has two versions, one with four answer choices per question and the other with five, and we use both for evaluation.
For MedMCQA, which does not have a public test set, we follow the approach taken by \citet{pmc-llama} and \citet{biomistral}, taking a random 80--20 train--validation split of the official training set and using the official validation set for testing.
For PubMedQA, we follow \citet{llm-clinical}, using the 211k artifically generated QA samples for training, and taking a 50--50 split on the 1k expert-labeled examples. For MMLU-Medical, we use the official split as provided. We provide the remaining dataset details in Appendix~\ref{sec:details-datasets}.

\paragraph{Visual QA Datasets.} For visual QA, we use VQA-RAD \citep{vqa-rad}, PathVQA \citep{pvqa}, SLAKE \citep{slake}, and MMMU-Medical \citep{mmmu} for evaluation. MMMU-Medical refers to a subset of MMMU corresponding to 5 subjects relevant to medicine: basic medical science, clinical medicine, diagnostics and laboratory medicine, pharmacy, and public health. For VQA-RAD, we address the train-test leakage and duplication issues in the official train--test splits, previously noted by \citet{med-flamingo}, by removing the training examples repeated in the test set and removing all duplicates in both sets. We then take a random 80--20 split on the training set to create a new train--validation split, as the official split does not include a validation set. For MMMU-Medical, which does not have a public test set, we randomly select 5 examples from the official validation set for validation, and reserve the remaining 25 examples for testing. For all other datasets, we use the official split as provided. We provide the remaining dataset details in Appendix~\ref{sec:details-datasets}.

\paragraph{Evaluation Metric.} Since we focus on closed-ended QA tasks, we use exact-match accuracy as our main evaluation metric. Following the Holistic Evaluation of Language Models (HELM) benchmark \citep{helm}, when we consider greedy decoding, we treat the text generated by a model (without any constraints on the vocabulary) to be its prediction, and check for an exact match between the prediction and the correct answer up to primitive string operations (e.g., lower-casing, removing white space/punctuation). To handle cases where the model simply repeats the list of answer choices or produces an ambiguous answer (e.g., selecting multiple answer choices), we take a conservative approach and treat the prediction to be incorrect, even if there is a match.
Meanwhile, to quantify the extent of \textit{improvement} from medical DAPT, we also consider the \textit{relative} accuracy of the medical model with respect to the general-domain model. Formally, we define relative exact-match accuracy as $\mathbb{E}[\mathds{1}[f_{\text{medical}}(x) = y] - \mathds{1}[f_{\text{general}}(x) = y]] \in [-1,1]$, where $f_{\text{medical}}$ and $f_{\text{general}}$ denote the medical and general-domain models, $x$ and $y$ denote the input prompt and answer in a QA pair from the test set, and $\mathds{1}[\cdot]$ denotes the indicator function. 
This metric quantifies the difference in accuracy between the medical model and the general-domain model.
To distinguish the two metrics, we refer to the former as the \textit{absolute} exact-match accuracy in subsequent discussions.

\begin{figure*}[t!]
    \centering
    \includegraphics[width=1\linewidth]{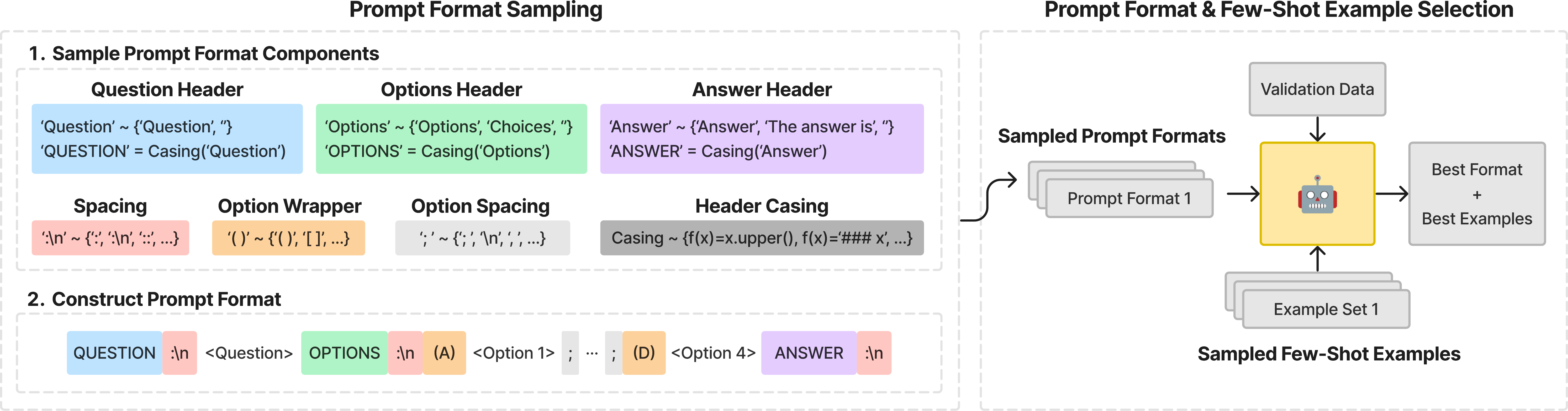}
    \caption{Overview of the prompt format sampling (left) and prompting strategy selection (right) process.}
    \label{fig:prompt-selection}
\end{figure*}

\paragraph{Assessing Statistical Significance.} Given the relatively small size of test datasets in medical QA benchmarks, it is important to assess whether the perceived improvements in performance from medical DAPT are attributable to chance.  To account for statistical uncertainty, we use the percentile bootstrap, re-sampling (with replacement) questions from the test set to get a sample of the same size as the original test set.  Within each resample, we compute the difference in accuracy for the paired models, and repeat this process for 10,000 iterations.  The resulting distribution of relative accuracy is used to derive a 95\% confidence interval, and we judge a difference to be statistically significant if this interval does not cross zero.  We do not perform any type of multiple-testing correction, which would have the effect of lowering the number of comparisons deemed to be significant.

\subsection{Zero-/Few-shot Prompting with Model-Specific Prompt Selection}\label{sec:prompting}

In this section, we provide an overview of our approach to assess whether medical DAPT leads to statistically significant improvements in zero-/few-shot medical QA performance. 
For few-shot prompting, we consider the 3-shot setting to ensure that the input prompt is shorter than the context window sizes for all models evaluated.
For evaluation, we pay special attention to two aspects. 
First, language models are highly sensitive to the choice of prompting strategy (e.g., prompt format, choice of few-shot examples), where seemingly insignificant changes to the prompt can lead to idiosyncratic model behavior \citep{how-can-we-know-what-lms-know,calibrate-before-use}. 
Second, prior works show that the ``optimal'' choice of prompt format rarely correlates between different models \citep{quantifying-lm-prompt-design}, suggesting that using a single, fixed prompt for all models for comparison can result in misleading conclusions.

To ensure a fair comparison that isolates the impact of medical DAPT, we treat the choice of prompt format and few-shot examples as additional hyperparameters when generating predictions, and tailor them to each model \textit{independently} (Figure \ref{fig:prompt-selection}). 
We first randomly sample 10 plausible prompt formats from a predefined search space and 10 different sets of few-shot examples from the training set of each dataset. We then search over all pairs of prompt formats (plus one additional manually designed default format) and few-shot examples, and select the best pair out of $(10+1) \times 10 = 110$ that results in the highest validation exact-match accuracy.
Given that a grid search at this scale can be computationally expensive, especially for datasets like MedMCQA that contain 37k validation QA pairs (see Table \ref{tab:datasets}), we randomly subsample 500 validation QA pairs for datasets that have more than 500. Using the vLLM framework \citep{vllm} for sampling model outputs, this leads to a runtime of around 5--15 minutes per trial, on 4 NVIDIA A6000 GPUs for the 70B models and 2 GPUs for the others.
We then generate predictions on the test set using the selected prompt format and few-shot samples. 
In the zero-shot setting, we only search over the prompt formats. 

\begin{figure*}[t!]
    \centering
    \begin{tabular}{@{}c@{}c@{}}
        \multicolumn{2}{c}{
            \begin{subfigure}{0.98\linewidth}
                \includegraphics[width=\linewidth]{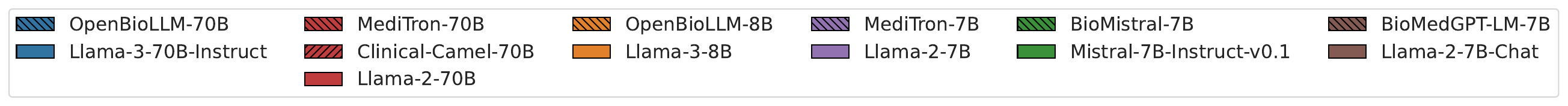}
            \end{subfigure}
        }
        \\
        \begin{subfigure}{0\linewidth}
        \end{subfigure} &
        \begin{subfigure}{0.98\linewidth}
            \includegraphics[width=\linewidth]{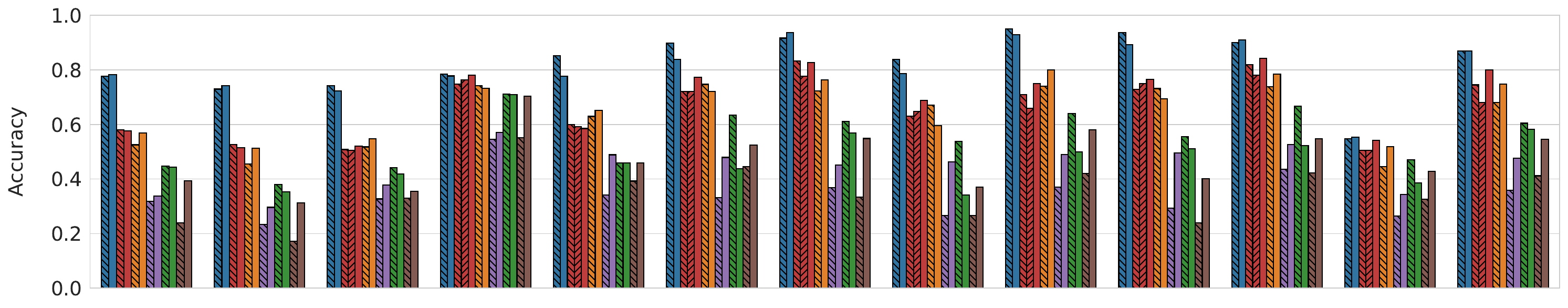}
        \end{subfigure}
        \\
        \begin{subfigure}{0\linewidth}
        \end{subfigure} &
        \begin{subfigure}{0.98\linewidth}
            \includegraphics[width=\linewidth]{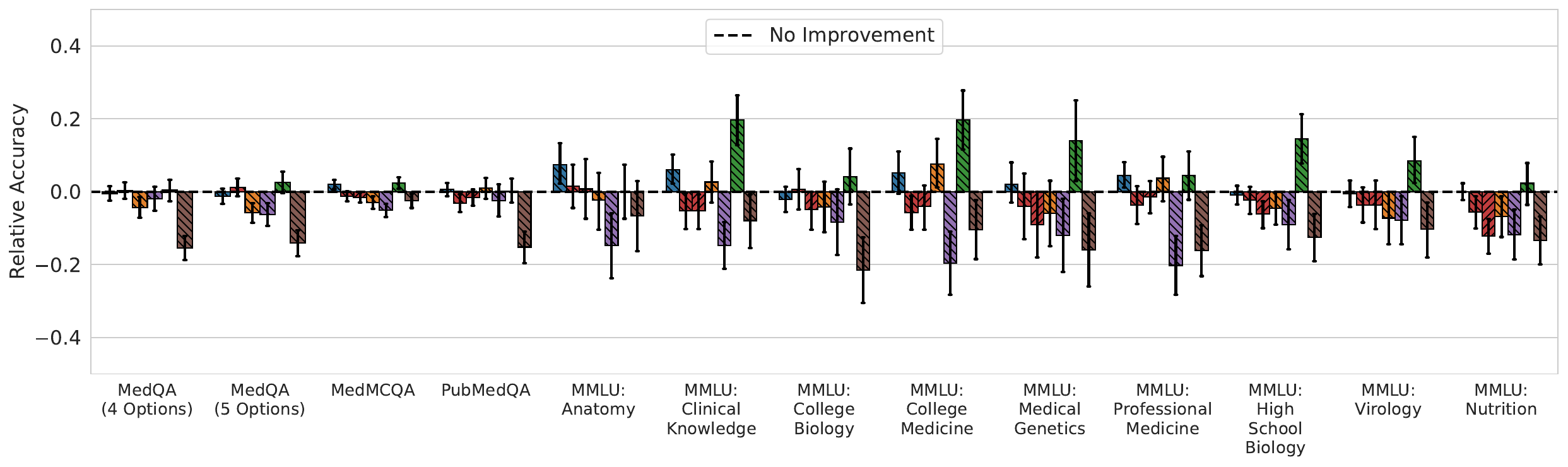}
        \end{subfigure}
    \end{tabular}
    \vspace{-10pt}
    \caption{Medical LLMs do not consistently show a statistically significant improvement over their general-domain counterparts in the 3-shot setting, after independently selecting the best prompt format and examples for each model. Top row shows the absolute exact-match accuracies on the test set, and bottom row shows the relative exact-match accuracies along with 95\% confidence intervals derived via bootstrapping on the test set (see Section~\ref{sec:eval-setup}). Here, we show the results for greedy decoding. 
    The 3-shot results for constrained decoding are similar (see Figure~\ref{fig:llm-logprob-acc-ci}(b)).}
    \label{fig:llm-acc-ci}
\end{figure*}

To define the prompt format search space, we follow the approach by \citet{quantifying-lm-prompt-design} and construct a context-free grammar of semantically equivalent yet syntactically distinct prompt formats (Figure~\ref{fig:prompt-selection}, left).
For the medical models that have a specific prompt format designed and recommended for closed-ended QA tasks (e.g., \textsc{BioMistral} \citep{biomistral}), we fix the prompt format to what is provided and only search over the choice of few-shot examples. In the case when such information is missing or only partially available (see Table~\ref{tab:prompting-details}), we search over both the prompt formats and few-shot examples. For instruction-tuned models, which typically have a structured conversational format (e.g., `\texttt{\#\#\# User:\ldots \#\#\# Assistant:\ldots}'') that is expected, we use the sampled question and answer templates to format each ``user'' query and ``assistant'' response. We provide the remaining details in Appendix~\ref{sec:details-prompting-selection}--\ref{sec:details-prompting}.

%% file: latex/sections/4_results.tex
Here, we summarize the main findings from the zero-/few-shot prompting  
experiments outlined in Section \ref{sec:eval-setup}.
Unless specified otherwise, we focus on the greedy decoding results in subsequent discussions and include the results for constrained decoding in Appendix \ref{sec:constrained-decoding}.
Overall, we find that all medical VLMs and nearly all medical LLMs fail to consistently improve over their general-domain counterparts in the zero-shot and few-shot prompting regimes.
Moreover, we demonstrate the importance of rigorous experimental design in surfacing this finding---performing pairwise model comparison with a single, fixed prompt optimized only for the medical model, while ignoring statistical uncertainty, paints a misleadingly optimistic picture of medical DAPT performance.

\begin{figure*}[t!]
    \centering
    \begin{tabular}{@{}c@{}c@{\hskip 2pt}c@{}c@{}}
        \multicolumn{4}{c}{
            \begin{subfigure}{0.8\linewidth}
                \includegraphics[width=\linewidth]{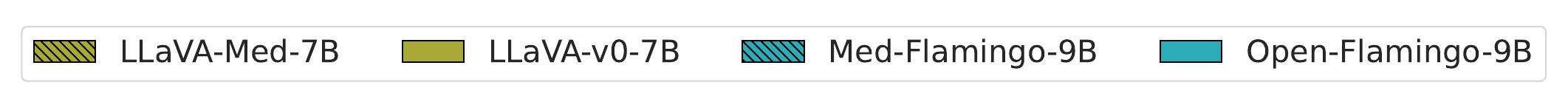}
            \end{subfigure}
        }
        \\
        \begin{subfigure}{0.03\linewidth}
            \makebox[\linewidth]{\raisebox{60pt}{{(a)}}}
        \end{subfigure} &
        \begin{subfigure}{0.47\linewidth}
            \includegraphics[width=\linewidth]{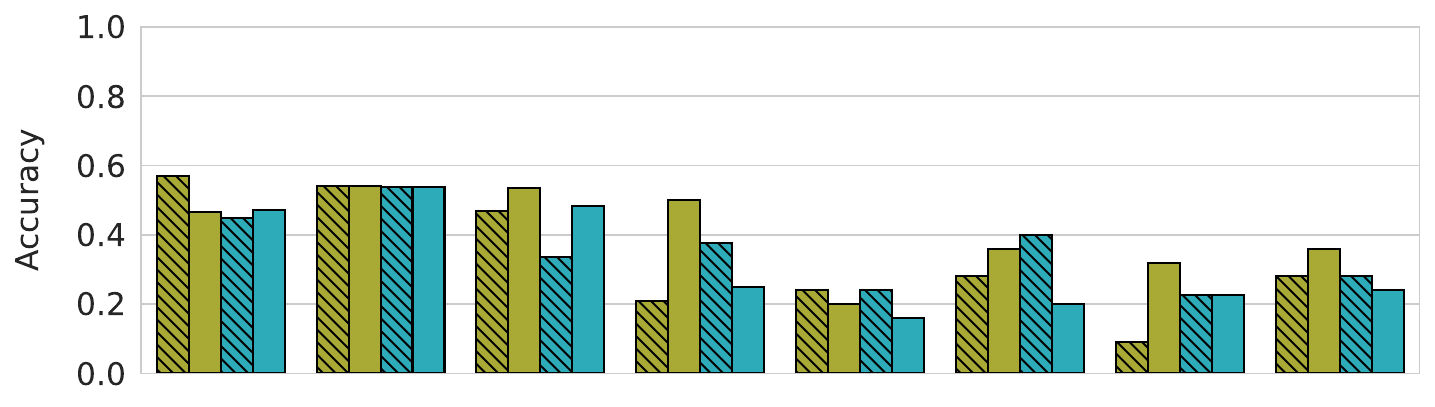}
        \end{subfigure} &
        \begin{subfigure}{0.03\linewidth}
            \makebox[\linewidth]{\raisebox{60pt}{{(b)}}}
        \end{subfigure} &
        \begin{subfigure}{0.47\linewidth}
            \includegraphics[width=\linewidth]{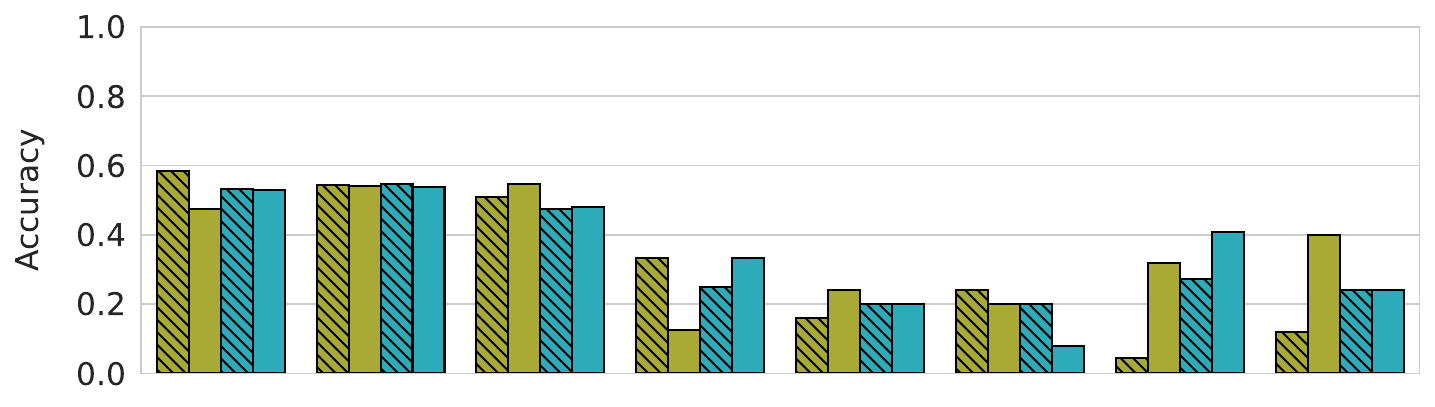}
        \end{subfigure}
        \\
        \begin{subfigure}{0.03\linewidth}
        \end{subfigure} &
        \begin{subfigure}{0.47\linewidth}
            \includegraphics[width=\linewidth]{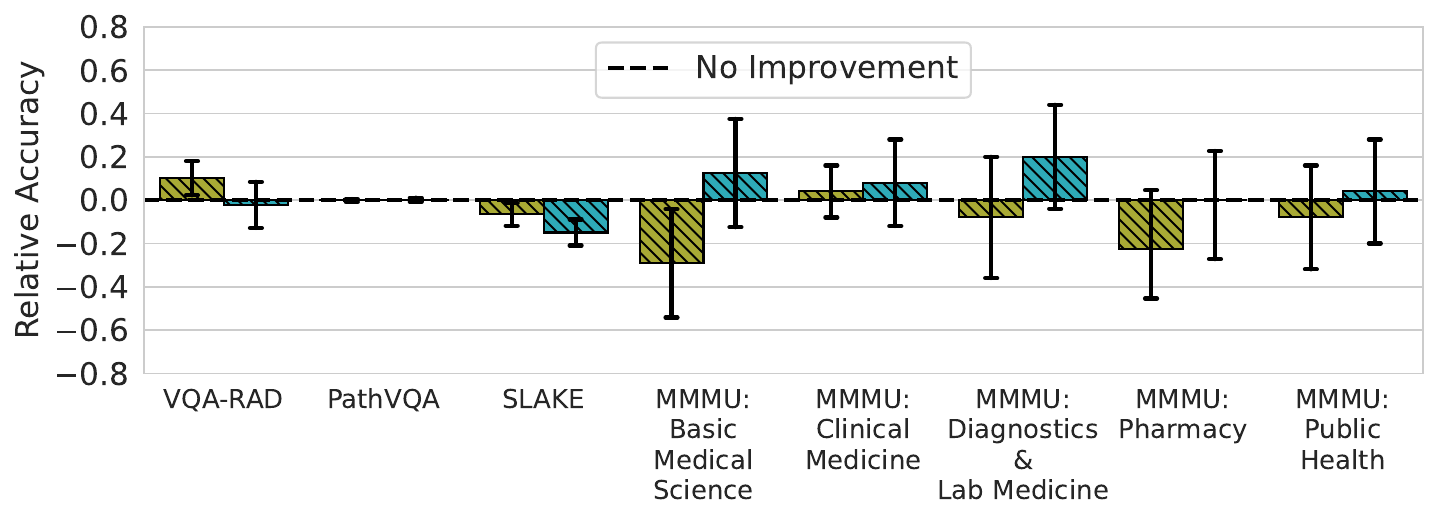}
        \end{subfigure} &
        \begin{subfigure}{0.03\linewidth}
        \end{subfigure} &
        \begin{subfigure}{0.47\linewidth}
            \includegraphics[width=\linewidth]{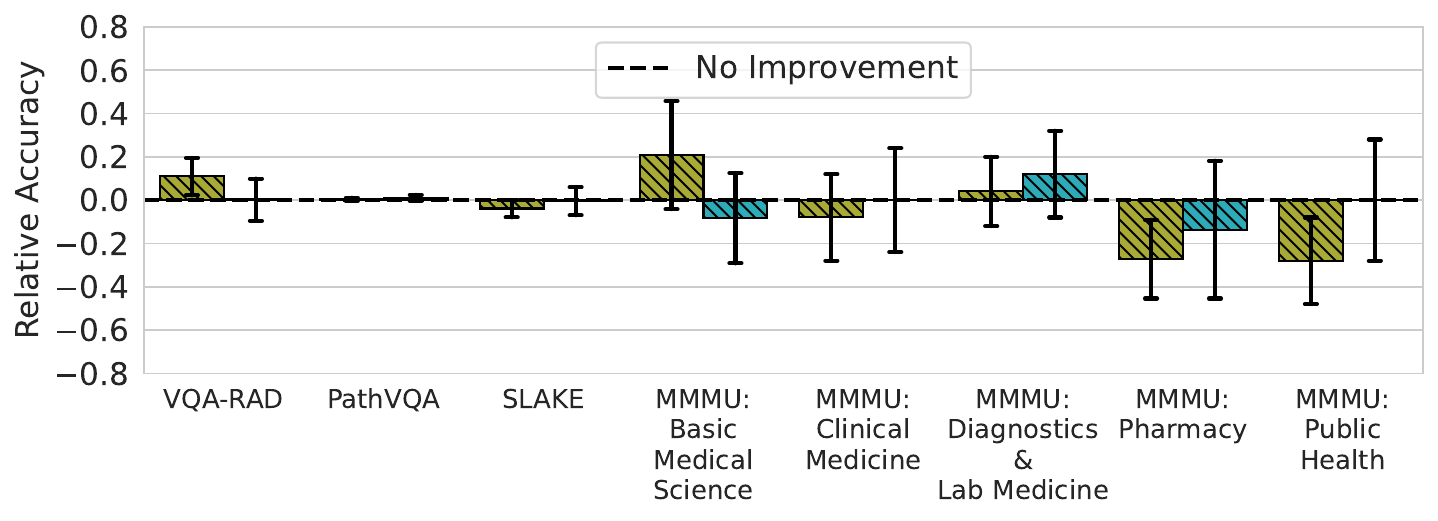}
        \end{subfigure}
    \end{tabular}
    \vspace{-7.5pt}
    \caption{Medical VLMs do not show a statistically significant improvement over their general-domain counterparts in the (a) zero-shot and (b) 3-shot settings, after independently selecting the best prompt format and examples for each model. Top row shows the absolute exact-match accuracies on the test set, and bottom row shows the relative exact-match accuracies along with 95\% confidence intervals derived via bootstrapping on the test set (see Section~\ref{sec:eval-setup}). Here, we show the results for greedy decoding. The results for constrained decoding are similar (see Figure~\ref{fig:vlm-logprob-acc-ci}).}
    \label{fig:vlm-acc-ci}
\end{figure*}

\paragraph{Finding 1: After model-specific prompt selection, the vast majority of medical models fail to consistently show a statistically significant improvement over the general-domain models.} 
In Figures~\ref{fig:llm-acc-ci}--\ref{fig:vlm-acc-ci}, we show the absolute and relative exact-match accuracies achieved by the medical and general-domain LLMs and VLMs in the zero-/few-shot prompting regime. 
For LLMs, we only show the 3-shot prompting results in the main text (see Appendix~\ref{sec:prompting-results-appendix} for results in the zero-shot setting, which are similar).
We exclude the results for \textsc{Clinical-Camel-70B} on both versions of MedQA, as the model has already been trained on a subset of the official training split (see Table 1 in \citet{clinical-camel}).
For VLMs, we show both zero-shot and 3-shot results, as \textsc{LLaVA-v0-7B} and \textsc{LLaVA-Med-7B} were not pretrained to handle inputs with multiple images. 
We calculate the confidence intervals via bootstrapping on the test set, as described in Section~\ref{sec:eval-setup}.

The top row of Figure~\ref{fig:llm-acc-ci} shows that the absolute exact-match accuracies are mostly similar between each model pair across all datasets and model scales, with marginal performance improvements.
In fact, the bottom row of Figure~\ref{fig:llm-acc-ci} shows that \textbf{only 2 out of 7 medical LLMs}---\textsc{OpenBioLLM-70B} and \textsc{BioMistral-7B}---show statistically significant improvements in performance, with the 95\% confidence intervals crossing zero relative accuracy in most cases for the other models.
When compared against their corresponding base models, \textsc{OpenBioLLM-70B} achieves a win rate of 30.8\%, tie rate of 69.2\%, and loss rate of 0\%, while \textsc{BioMistral-7B} achieves a win rate of 46.2\%, tie rate of 53.8\%, and loss rate of 0\% (Table~\ref{tab:win-tie-loss-rates-3}).
Notably, \textsc{MediTron-7B} and \textsc{BioMedGPT-LM-7B} actually show significantly worse performance than their base models, with loss rates of 76.9\% and 92.3\%, respectively.
Similar trends hold for the zero-shot setting (Figure \ref{fig:llm-0-acc-ci} and Table \ref{tab:win-tie-loss-rates-0}), where only \textsc{Clinical-Camel-70B} and \textsc{BioMistral-7B} show statistically significant improvements. 

We note that, while \textsc{OpenBioLLM-70B} shows improvement in the 3-shot setting, it does not show improvement in the zero-shot setting (winning on 7.7\% and losing on 23.1\% of tasks, see Table~\ref{tab:win-tie-loss-rates-0}), and vice versa for \textsc{Clinical-Camel-70B} (winning on 0\% of tasks and losing on 36.4\% of tasks in the 3-shot setting, see Table~\ref{tab:win-tie-loss-rates-3}), \textbf{leaving \textsc{BioMistral-7B} as the only medical LLM that wins more than it loses} against its base model (\textsc{Mistral-7B-Instruct-v0.1}) in both settings, albeit with relatively low absolute performance.

In Figure~\ref{fig:vlm-acc-ci}, we make similar observations for medical VLMs in both zero-shot and 3-shot settings, where both \textsc{LLaVA-Med-7B} and \textsc{Med-Flamingo-9B} are \textbf{virtually indistinguishable from their base models in terms of performance, showing no statistically significant improvements}.
Tables~\ref{tab:win-tie-loss-rates-0}--\ref{tab:win-tie-loss-rates-3} show that \textsc{LLaVA-Med-7B} achieves win/tie/loss rates of 12.5\%/62.5\%/25.0\% in both zero-shot and 3-shot settings, while \textsc{Med-Flamingo-9B} achieves win/tie/loss rates of 0\%/87.5\%/12.5\% in the zero-shot setting and 0\%/100\%/0\% in the 3-shot setting.
Meanwhile, we note that the confidence intervals for the MMMU-Medical datasets tend to be much wider than for the other visual QA datasets, as the test sets only include 25 QA examples for each subject (Table~\ref{tab:datasets}).

We similarly observe limited improvements overall with constrained decoding (see Appendix~\ref{sec:constrained-decoding-finding1}). As shown in Figure~\ref{fig:opt-logprob-ci-acc}(a), when we aggregate the results over all (model pair, QA dataset) combinations, medical LLMs achieve win/tie/loss rates of 16.9\%/68.6\%/14.5\% in the zero-shot setting and 11.2\%/74.1\%/14.7\% in the 3-shot setting, while medical VLMs achieve win/tie/loss rates of 6.3\%/87.5\%/6.3\% in the zero-shot setting and 0\%/93.8\%/6.3\% in the 3-shot setting. In fact, no medical VLM shows improvement over its base model regardless of the decoding strategy. Meanwhile, as shown in Tables~\ref{tab:win-tie-loss-rates-0-logprob}--\ref{tab:win-tie-loss-rates-3-logprob}, we find that some medical LLMs show larger improvements with constrained decoding (notably, \textsc{MediTron-70B} and \textsc{MediTron-7B}), although the results are mixed (e.g., \textsc{Clinical-Camel-70B} performs worse in the zero-shot setting with constrained decoding).

\textit{In summary, these results suggest that when prompted with the ``right'' set of examples in an appropriate format, general-domain models may already exhibit the capacity to achieve performance competitive with medically adapted models, on various medical QA tasks.} 

\begin{figure*}[t!]
    \centering
    \includegraphics[width=0.95\linewidth]{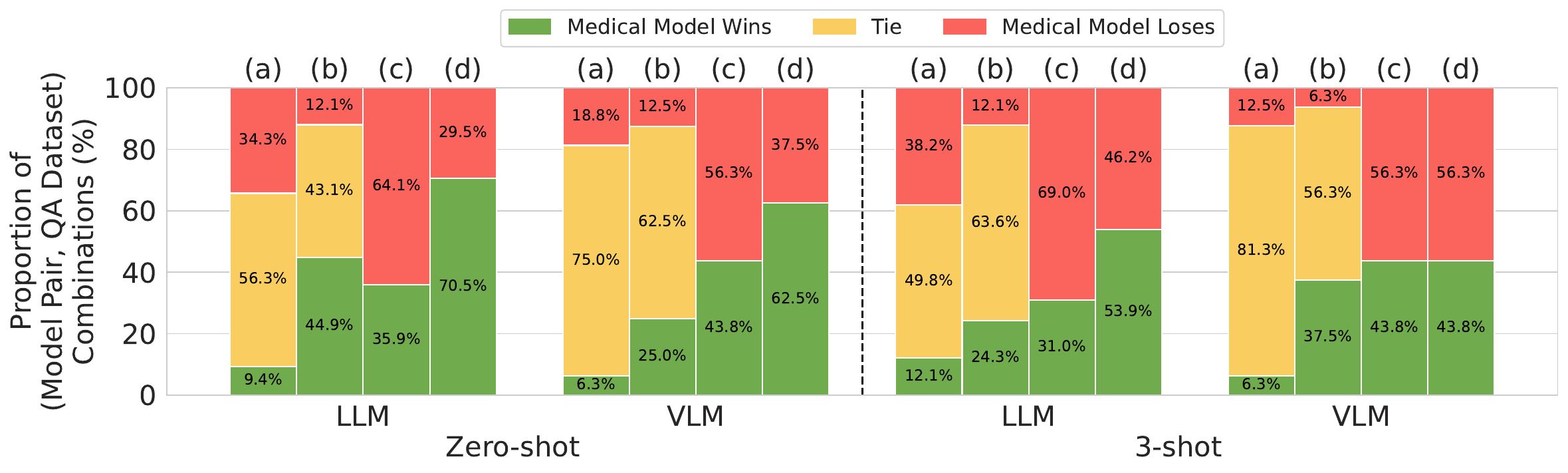}
    \vspace{-5pt}
    \caption{Optimizing the prompt for only the medical model and comparing models without accounting for statistical uncertainty can overestimate the performance improvements from medical DAPT. We show the win/tie/loss rate (\%) of medical models vs. their base models across all (model pair, QA dataset) combinations, when (a) independently optimizing the prompt for each model and performing statistical testing, (b) optimizing the prompt only for the medical model and performing statistical testing, (c) independently optimizing the prompt for each model without statistical testing, and (d) optimizing the prompt only for the medical model without statistical testing. Here, we show the results for greedy decoding. The results for constrained decoding are similar (see Figure~\ref{fig:opt-logprob-ci-acc}).}
    \label{fig:opt-ci-acc}
\end{figure*}

\paragraph{Finding 2: Using a single, fixed prompt for all models and overlooking statistical uncertainty may overestimate the performance benefits of medical DAPT.} Based on Finding 1, we further investigate whether the conclusions differ if the same prompt is used for each pair of medical and general-domain models. In particular, we consider whether selecting a prompt only for the medical model, following Section \ref{sec:prompting}, and using it for the corresponding general-domain model can widen the performance gap between each pair. We also assess whether this gap becomes amplified when models are compared without accounting for statistical uncertainty, which is often done in practice.

In Figure~\ref{fig:opt-ci-acc}, we show how the win/tie/loss rates of the medical models, computed over all (model pair, QA dataset) combinations,
change as we vary the following aspects of the experimental setup:
\begin{enumerate}[topsep=5pt,itemsep=-0.5ex]
    \item select prompts for each model independently vs. only based on the medical model;
    \item determine a win for the medical model based on confidence intervals in relative accuracy vs. raw absolute accuracy.
\end{enumerate}
We note that when comparing each model pair based on absolute accuracy, there are no ties, as the real-valued absolute accuracies are rarely identical.
In Appendix~\ref{sec:prompting-results-appendix}, we include Figures~\ref{fig:llm-acc-ci-med}--\ref{fig:vlm-acc-ci-med} to show how the absolute and relative exact-match accuracies change when the prompt is only optimized for the medical model. We also include Tables~\ref{tab:win-tie-loss-rates-med-0}--\ref{tab:win-tie-loss-rates-med-3} to show changes in win/tie/loss rates.
We show the same set of results for constrained decoding in Figures~\ref{fig:llm-acc-ci-med-logprob}--\ref{fig:vlm-acc-ci-med-logprob} and Tables~\ref{tab:win-tie-loss-rates-med-0-logprob}--\ref{tab:win-tie-loss-rates-med-3-logprob} in Appendix~\ref{sec:constrained-decoding}.

Overall, we find that for both LLMs and VLMs, the performance improvement from using a medically adapted model instead of its general-domain counterpart can be substantially overestimated when (i) the prompt is only tailored to the medical model; and (ii) the models are compared only based on their absolute accuracies. 
Notably, in the zero-shot setting, the win rate increases from 9.4\% to 70.5\% for medical LLMs and from 6.3\% to 62.5\% for medical VLMs, 
when only performing prompt selection for the medical model and comparing based on absolute accuracy.
Figure~\ref{fig:opt-logprob-ci-acc} in Appendix~\ref{sec:constrained-decoding-finding2} shows a similar trend in the win/tie/loss rates, when the model predictions are generated via constrained decoding.
\textit{These results highlight the importance of accounting for LLM/VLM sensitivity to the prompting details, as suggested by \citet{quantifying-lm-prompt-design}, and the statistical uncertainty in model comparison, in order to draw reliable conclusions about the effectiveness of medical DAPT.}

%% file: latex/sections/5_discussion.tex
In this work, we investigated the effectiveness of DAPT 
for training medically specialized LLMs and autoregressive VLMs 
suitable for knowledge-intensive medical (visual) QA tasks.
To that end, we compared several pairs of state-of-the-art medical LLMs/VLMs 
to their general-domain counterparts,
whose only differences lie in medical DAPT 
and are exactly identical in model architecture and scale.
Our work diverges from prior works 
by providing a direct apples-to-apples comparison 
of medical and general-domain models 
while accounting for LLM/VLM sensitivity to prompting details
and assessing the statistical significance of the results.

Across both model classes and all model scales, 
we found that the performance benefits from medical DAPT 
largely disappear when we 
(i) tailor the prompt format and choice of few-shot examples 
to each medical and general-domain model separately;
and (ii) account for statistical uncertainty in model comparison.
In particular, we found that when we optimize the prompt only for the medical model 
and compare each model pair based on their absolute accuracies without accounting for uncertainty, 
the performance improvements from medical DAPT can be overestimated, 
potentially leading to unreliable conclusions about the benefits of medical DAPT.
For example, in the zero-shot setting,
evaluation under this setup leads to the conclusion that medical LLMs and VLMs, 
on average, outperform the corresponding general-domain models in 70.5\% and 62.5\% of all QA tasks, 
while the improvements are in reality statistically significant in only 9.4\% and 6.3\% of tasks 
after optimizing the prompt for each model to ensure a fair comparison.

Our findings suggest that for state-of-the-art general-domain LLMs and VLMs, 
the performance benefits from additionally pretraining 
on medical data from public sources such as PubMed may be limited. 
Notably, almost all of the medical models used in our evaluation 
use PubMed as the primary source of pretraining data 
for medical adaptation (Table~\ref{tab:models}), 
while open-source datasets commonly used 
for pretraining the general-domain base models in the first place 
(e.g., the Pile \citep{the-pile}, S2ORC \citep{s2orc}) 
often already include PubMed data. 
Prior works also suggest that the intrinsic \textit{capacity} of LLMs to solve a downstream task is largely obtained during the \textit{initial} pretraining phase, and that post-training adjustments and prompt engineering efforts may only help elicit the existing capabilities \citep{prompt-programming,rethinking-demo}.
Thus, we argue that any claims about improvement 
from a proposed medical DAPT procedure 
should be evidenced by rigorous head-to-head comparisons 
against the corresponding general-domain model, 
in order to draw reliable conclusions about its effectiveness.

%% file: latex/sections/6_limitations.tex
We discuss our findings with the following caveats.

First, there is a vast and growing set of papers on applying medical DAPT to various general-domain base models, and we could not hope to compare all publicly available models here. 
While we selected the models to cover a wide range of general-domain base models and model scales (7B--70B) (Table~\ref{tab:models}) and included some of the latest models (e.g., \textsc{OpenBioLLM} and \textsc{Llama-3}), 
it is always possible that some newly released models do in fact yield better zero- or few-shot performance on medical QA.

Second, we focus in this paper on the narrower task of closed-ended medical QA.  In part, this choice reflects the fact that such benchmarks are well-standardized and highly publicized. However, they do not reflect the breadth of possible applications of LLMs and VLMs in medical domains. For instance, \citet{med-palm-2} show that medical LLMs such as \textsc{Med-PaLM-2} can produce physician-level answers to open-ended consumer health queries, 
and \citet{llms-are-clinical-info} demonstrate the potential of using LLMs for extracting information from structured clinical notes.
Some would argue that such tasks are a more realistic application of such models in practice, and it is certainly possible that an analysis like ours would find improved performance on such tasks, though we do not investigate these tasks in the present work.

Third, we do not consider downstream fine-tuning of models subject to medical DAPT. In part, this reflects issues of computational cost (e.g., to fine-tune 70B-parameter models) and the added complexity of reproducing a fine-tuning procedure, versus using publicly available model checkpoints. However, we acknowledge that zero- and few-shot performance are only part of a broader narrative around the claimed benefits of medical DAPT, which generally includes the additional claim that it provides a better initialization for downstream fine-tuning \citep{meditron,llava-med}.

While we acknowledge the limitations above, we do not believe they detract from the value of this work.  We hope that our results call attention to a need for rigorous head-to-head evaluations when making similar claims of improved performance via medical DAPT, whether with other models, on other clinical tasks, or with respect to fine-tuning versus zero-/few-shot performance.

%% file: latex/sections/appendix_datasets.tex
\begin{table}[h!]
    \centering
    \caption{Summary of the number of examples in the train, validation, and test sets of all textual and visual QA datasets used for evaluation, in the top and bottom sections, respectively.}
    \label{tab:datasets}
    \resizebox{\linewidth}{!}{
    
    \begin{tabular}{@{}l@{\hskip 7pt}c@{\hskip 7pt}c@{\hskip 7pt}c@{\hskip 3pt}}
        \toprule
        Dataset & Train & Validation & Test \\
        \midrule
        MedQA (4 \& 5 Options) & 10178 & 1272 & 1273 \\
        MedMCQA & 146257 & 36565 & 4183 \\
        PubMedQA & 211269 & 500 & 500 \\
        MMLU: Anatomy & 5 & 14 & 135 \\
        MMLU: Clinical Knowledge & 5 & 29 & 265 \\
        MMLU: College Biology & 5 & 16 & 144 \\
        MMLU: College Medicine & 5 & 22 & 173 \\
        MMLU: High School Biology & 5 & 32 & 310 \\
        MMLU: Medical Genetics & 5 & 11 & 100 \\
        MMLU: Nutrition & 5 & 33 & 306 \\
        MMLU: Professional Medicine & 5 & 31 & 272 \\
        MMLU: Virology & 5 & 18 & 166 \\
        \midrule
        VQA-RAD & 820 & 205 & 272 \\
        PathVQA & 9806 & 3135 & 3391 \\
        SLAKE & 1943 & 422 & 415 \\
        MMMU: Basic Medical Science & 5 & 5 & 25 \\
        MMMU: Clinical Medicine & 5 & 5 & 25 \\
        MMMU: Diag. \& Lab Medicine & 5 & 5 & 25 \\
        MMMU: Pharmacy & 5 & 5 & 25 \\
        MMMU: Public Health & 5 & 5 & 25 \\
        \bottomrule
    \end{tabular}
    }
\end{table}

\noindent In this section, we provide additional details on the textual and visual QA datasets introduced in Section \ref{sec:eval-setup}. In Table \ref{tab:datasets}, we summarize the number of QA examples included in the train, validation, and test sets of each dataset, after following the preprocessing steps detailed in Section \ref{sec:eval-setup}.
For VQA-RAD \citep{vqa-rad}, PathVQA \citep{pvqa}, and SLAKE \citep{slake}, we only show the number of \textit{closed-ended} visual QA examples, since our evaluations focus on closed-ended visual QA. 
For the datasets that required additional splits from the official train-validation-test split (e.g., due to the lack of a public test set), we include all of the fixed random seeds in our repository for reproducibility.

%% file: latex/sections/appendix_prompt-selection.tex
In this section, we provide additional details on how we define the prompt format search space discussed in Section \ref{sec:prompting}. 
We construct a context-free grammar of plausible prompt formats following the approach by \citet{quantifying-lm-prompt-design} (see Section 3.1 and Appendix A of the paper for reference). Using the Backus-Naur notation, we first define the basic fields $H_q$ for the question header (e.g., \texttt{``\#\#\# Question:''}), $H_c$ for the answer choice header ('e.g., \texttt{``\#\#\# Options:''}), and $H_a$ for the answer header (e.g., \texttt{``\#\#\# Answer:''}) as
\begin{align}
    H_q(f_{\text{case}}, d_q, s_1) &::= f_{\text{case}}(d_q) s_1 \langle\text{text}\rangle, \nonumber \\
    H_c(f_{\text{case}}, d_c, s_1) &::= f_{\text{case}}(d_c) s_1, \nonumber \\
    H_a(f_{\text{case}}, d_a, s_1) &::= f_{\text{case}}(d_a) s_1 \langle\text{text}\rangle, \nonumber
\end{align}
where $f_\text{case} \in \mathcal{F}_{\text{case}}$ denotes the casing function (e.g., \texttt{x} $\mapsto$ ``\#\#\# '' + \texttt{x}, \texttt{x} $\mapsto$ \texttt{x.upper()}), $d_q \in D_q$ denotes the question descriptor (e.g., \texttt{``Question''}), $d_c \in D_c$ denotes the answer choice descriptor (e.g., \texttt{``Options''}), $d_a \in D_a$ denotes the answer descriptor (e.g., \texttt{``Answer''}), $s_1 \in S_1$ denotes the header separator (e.g., \texttt{`:'}), and $\langle\text{text}\rangle$ denotes a text placeholder. 
For formatting the list of answer choices, we also define the basic fields $C$ for formatting each answer choice (e.g., \texttt{``(A) yes''}) and $L$ for the concatenation of all answer choices as follows:
\begin{align}
    C(f_{\text{wrap}},f_{\text{index}}, i) &::= f_{\text{wrap}}(f_{\text{index}}(i))\langle\text{text}\rangle, \nonumber \\
    L(f_{\text{wrap}},f_{\text{index}},n,s_2) &::= C(f_{\text{wrap}},f_{\text{index}},0) s_2 \ldots \nonumber \\
    &\quad s_2 C(f_{\text{wrap}},f_{\text{index}},n-1), \nonumber
\end{align}
where $f_{\text{wrap}} \in \mathcal{F}_{\text{wrap}}$ denotes the wrapper function for the answer choice letter (e.g., \texttt{x} $\mapsto$ \texttt{``('' + x + ``)''}), $f_{\text{index}} \in \mathcal{F}_{\text{index}}$ denotes the numbering function that converts an integer index into a number format (e.g., 0 $\rightarrow$ ``A''), $i \in \mathbb{Z}^+$ denotes the index of a particular answer choice from the list, $s_2 \in S_2$ denotes the answer choice separator, $n$ denotes the number of answer choices, and $\langle$text$\rangle$ denotes a text placeholder. The full prompt format $P(f_{\text{case}},f_{\text{wrap}},f_{\text{index}},d_q,d_c,d_a,s_1,s_2,n)$ is then constructed by concatenating all of the headers and the answer choices, while adding space $t \in T$ (e.g., \texttt{``\textbackslash n''}) in-between:
\begin{align}\label{eq:prompt-format}
    P &::= H_q t H_c t L t H_a,
\end{align}
where we have left out the notations for the arguments for notational simplicity.

To define the prompt format search space, we instantiate the grammar above with the descriptors, separators, spaces, and functions shown below.\\

\noindent \textbf{Descriptors:}
\begin{align}
    D_q &= \{\texttt{``Question''}, \texttt{``''}\}; \nonumber \\
    D_c &= \{\texttt{``Options''}, \texttt{``Choices''}, \texttt{``''}\}; \nonumber \\
    D_a &= \{\texttt{``Answer''}, \texttt{``The answer is''}\}. \nonumber
\end{align}

\noindent \textbf{Separators:}
\begin{align}
    S_1 &= \{\texttt{``: '', `` : '', `` :: '', ``:\textbackslash n'', ``= ''}, \nonumber \\
    &\quad\;\;\; \texttt{`` = '', `` == '', ``=\textbackslash n'', `` - '',} \nonumber \\
    &\quad\;\;\; \texttt{`` -- '', ``---'', ``\textbackslash n'', ``\textbackslash n\textbackslash n''}\}; \nonumber \\
    S_2 &= \{\texttt{``\textbackslash n'', ``\textbackslash \textbackslash'', ``; '', `` || '', `` ''} \nonumber \\
    &\quad\;\;\; \texttt{``;\textbackslash n'', ``;\textbackslash n\textbackslash n'', ``, ''}\}. \nonumber
\end{align}

\noindent \textbf{Spaces:}
\begin{align}
    T &= \{\texttt{``\textbackslash n'', ``\textbackslash n\textbackslash n'', `` || '', `` ''}\}. \nonumber
\end{align}

\noindent \textbf{Casing, Wrapper, and Numbering Functions:}
\begin{align}
    \mathcal{F}_{\text{case}} &= \{\texttt{x} \mapsto \texttt{x}, \texttt{x} \mapsto \texttt{x.title()}, \nonumber \\
    &\quad\;\;\; \texttt{x} \mapsto \texttt{x.upper()}, \texttt{x} \mapsto \texttt{x.lower()} \nonumber \\
    &\quad\;\;\; \texttt{x} \mapsto \texttt{``\#\#\# '' + x} \nonumber \\
    &\quad\;\;\; \texttt{x} \mapsto \texttt{``**'' + x + ``**''} \}; \nonumber \\
    \mathcal{F}_{\text{wrap}} &= \{\texttt{x} \mapsto \texttt{``('' + x + ``)''}, \texttt{x} \mapsto \texttt{ x + ``.''} \nonumber \\
    &\quad\;\;\; \texttt{x} \mapsto \texttt{ x + ``)''},  \texttt{x} \mapsto \texttt{``['' + x + ``]''} \nonumber \\
    &\quad\;\;\; \texttt{x} \mapsto \texttt{ x + `` )''}, \texttt{x} \mapsto \texttt{``<'' + x + ``>''}\}; \nonumber \\
    \mathcal{F}_{\text{index}} &= \{\texttt{x} \mapsto \texttt{chr(ord(``A'') + x)} \}. \nonumber
\end{align}
To randomly sample a prompt format accepted by the grammar, we randomly sample each of these components and construct the full prompt format, following Equation~\eqref{eq:prompt-format}. Below, we show an example QA pair from the MedQA dataset (four answer choices), formatted according to the formats sampled from the prompt format space defined by the above context-free grammar.\\

\noindent \textbf{Example 1:}
\begin{mdframed}
A key factor facilitating the application of nested case-control studies from the MACS was:

\noindent OPTIONS -- A ) Data collection

\noindent B ) Establishment of a repository of biologic specimens

\noindent C ) Participant interest

\noindent D ) Administration of the questionnaire by staff

\noindent THE ANSWER IS -- B ) Establishment of a repository of biologic specimens
\end{mdframed}

\noindent \textbf{Example 2:}
\begin{mdframed}
QUESTION -- A key factor facilitating the application of nested case-control studies from the MACS was:

\noindent CHOICES -- [A] Data collection; [B] Establishment of a repository of biologic specimens; [C] Participant interest; [D] Administration of the questionnaire by staff

\noindent ANSWER -- [B] Establishment of a repository of biologic specimens
\end{mdframed}

%% file: latex/sections/appendix_prompting.tex
In this section, we summarize the prompting details made available for the medical LLMs and VLMs used in our evaluation (Appendix \ref{sec:reproducibility}), and the default prompt formats used for each LLM (Appendix \ref{sec:llm-prompt-templates}) and VLM (Appendix \ref{sec:vlm-prompt-templates}), which have been reproduced based on the former.

\begin{table*}[ht!]
    \centering
    \caption{Summary of all of the prompting details made available for each medical LLM and VLM used for evaluation. For each column, a checkmark ({\greencheckmark}) indicates that the information was fully provided, a triangle ({\orangetriangle}) indicates that the information was partially provided (e.g., random sampling without information about the seeds), and a cross ({\redx}) indicates that the information was not provided at all. ``N/A'' indicates that the corresponding information is not available due to its irrelevance to the evaluation setup considered in the paper (e.g., lack of few-shot example details because the model was only originally evaluated in zero-shot or fine-tuning regimes).}
    \label{tab:prompting-details}
    \resizebox{0.9\linewidth}{!}{
    \begin{tabular}{@{}l@{\hskip 7pt}c@{\hskip 7pt}c@{\hskip 7pt}c@{\hskip 7pt}c@{\hskip 7pt}c@{}}
        \toprule
        Model & \shortstack{System\\Prompt} & \shortstack{Zero-/Few-Shot\\Prompt Format} & \shortstack{Few-Shot\\Examples} & \shortstack{Sampling\\Details} \\
        \midrule
        \textsc{OpenBioLLM} \citep{OpenBioLLMs} & \greencheckmark & \orangetriangle & \redx & \redx \\
        \textsc{Clinical-Camel}~\citep{clinical-camel} & \redx & \orangetriangle & \redx & \orangetriangle \\
        \textsc{BioMistral} \citep{biomistral} & \greencheckmark & \greencheckmark & \redx & \orangetriangle \\
        \textsc{MediTron} \citep{meditron} & \greencheckmark & \orangetriangle & \greencheckmark & \greencheckmark \\
        \textsc{BioMedGPT-LM} \citep{biomedgpt} & \redx & \redx & N/A & \redx \\
        \midrule
        \textsc{LLaVA-Med} \citep{llava-med} & \orangetriangle & \orangetriangle & N/A & \orangetriangle \\
        \textsc{Med-Flamingo} \citep{med-flamingo} & \greencheckmark & \orangetriangle & \redx & \redx \\
        \bottomrule
    \end{tabular}
    }
\end{table*}

\subsection{Reproducibility of Prompting Details}\label{sec:reproducibility}
In Table \ref{tab:prompting-details}, we provide a summary of all of the prompting details available (in the context of closed-ended medical QA) for all medical LLMs and VLMs used in our evaluation. We share these details to demonstrate our best efforts with reproducing the original prompting setups considered for performing our evaluations. In particular, we focus on whether the following four components are explicitly made available, either in the original publications or the publicly released code repository: (i) system prompt; (ii) zero-/few-shot prompt format (used for closed-ended QA tasks); (iii) the choice of few-shot examples; and (iv) details on how the text generations are sampled (e.g., softmax temperature, top-$p$, beam size, random seeds used for sampling).
Below, we provide detailed clarifications for each model. 

\paragraph{\textsc{OpenBioLLM}~\citep{OpenBioLLMs}.} For the \textsc{OpenBioLLM} models, we follow the instructions provided in the model cards posted by the authors on HuggingFace, for the \href{https://huggingface.co/aaditya/Llama3-OpenBioLLM-70B}{70B-parameter} and \href{https://huggingface.co/aaditya/Llama3-OpenBioLLM-8B}{8B-parameter} models. We use the recommended system prompt and the \textsc{Llama-3}-based conversational prompt format. Meanwhile, in Table~\ref{tab:prompting-details}, we treat the prompt format as partially missing, as the exact format that was used to format each question (``user'' query) and answer (``assistant'' response) for evaluation on closed-ended multiple-choice questions is not provided. At the time of writing, there are no additional details about the models that have been publicly released, beyond what is provided in the model cards. We include the default prompt format used for \textsc{OpenBioLLM} in Appendix~\ref{sec:openbiollm-template}.

\paragraph{\textsc{Clinical-Camel}~\citep{clinical-camel}.} For \textsc{Clinical-Camel}, we use the conversational prompt format used in the official GitHub repository, which corresponds to the official chat format for \textsc{Llama-2}~\citep{llama-2}. 
As the system prompts and few-shot examples used for the main evaluations in the paper are not provided, we use our own manually designed default system prompt and search over different choices of few-shot examples.
For sampling, the \href{https://github.com/bowang-lab/clinical-camel/blob/main/evaluation/get_model_answer.py#L54}{evaluation code} uses default temperature setting of 0.7 (albeit without the random seeds), which differs from our evaluation setup.
We include the default prompt format used for \textsc{Clinical-Camel} in Appendix~\ref{sec:clinical-camel-template}.

\paragraph{\textsc{BioMistral}~\citep{biomistral}.} For \textsc{BioMistral}, we use the system prompt and zero-/few-shot prompt format provided in Appendix F of the paper. At the time of writing, the code repository is not publicly available, and the paper does not provide details on what few-shot examples were used for evaluation. In Section 4.3 of the paper, \citet{biomistral} mention that the output vocabulary is constrained to be one of the answer choices in lettered format (e.g., one of [A,B,C,D]) to force the model to avoid generating irrelevant tokens in its output. Meanwhile, it is not explicitly clear whether (i) the filtered token with the highest probability was treated as the model's prediction or (ii) a token was randomly sampled based on the renormalized token probabilities.
We also note that the vocabulary filtering procedure makes their evaluation setup different from ours, as we use greedy decoding to sample the model outputs without any constraints on the vocabulary (see Section \ref{sec:eval-setup}). We include the default prompt format used for \textsc{BioMistral} in Appendix~\ref{sec:biomistral-template}.

\paragraph{\textsc{MediTron} \citep{meditron}.} For the \textsc{MediTron} models, we use the system prompts---tailored specifically to MedQA, MedMCQA, PubMedQA, and the MMLU datasets---provided in Table 2 of the paper. For the prompt formats, we use the ones provided in the official GitHub repository, as the prompt formats (those with special \texttt{`<|im\_start|>'} and \texttt{`<|im\_end|>'} tokens, following the \href{https://github.com/openai/openai-python/blob/release-v0.28.0/chatml.md}{ChatML format}) shown in the paper are only applicable to the fine-tuned models (see \href{https://github.com/epfLLM/meditron/issues/13#issuecomment-1845955741}{this discussion} from the official GitHub repository). In particular, we refer to the prompt formats provided in the \href{https://github.com/epfLLM/meditron/blob/a7c7cda3014e537f0df2ec58f836fbe920d6283b/evaluation/benchmarks.py#L622}{dataset preprocessing} code and used for \href{https://github.com/epfLLM/meditron/blob/a7c7cda3014e537f0df2ec58f836fbe920d6283b/evaluation/inference.py#L188}{evaluation} to determine the default prompt format for both the 70B- and 7B-parameter models. However, we were unable to reliably reproduce the zero-/few-shot prompting performance using this prompt format, and therefore perform a grid search over the prompt formats as well for model-specific prompt selection. In the \href{https://github.com/epfLLM/meditron/blob/a7c7cda3014e537f0df2ec58f836fbe920d6283b/evaluation/inference.py#L188}{evaluation code}, \citet{meditron} provide the random seeds used for sampling the few-shot examples; however, we also search over the set of few-shot examples to consider a larger number of few-shot example choices. 
For sampling, we use the same greedy decoding approach as considered in the paper (referred to ``Top Token Selection'' in Section 4.3 of the paper).
We include the default prompt format used for \textsc{MediTron} in Appendix~\ref{sec:meditron-template}.

\paragraph{\textsc{BioMedGPT-LM}~\citep{biomedgpt}.} While \textsc{BioMedGPT-LM} was evaluated on textual medical QA datasets such as MedMCQA and PubMedQA, the evaluation was performed only in the supervised fine-tuning regime, and the prompt formats used for these datasets are not available, to the best of our knowledge.
Meanwhile, the official GitHub repository provides \href{https://github.com/PharMolix/OpenBioMed/blob/main/examples/biomedgpt_inference.ipynb}{Jupyter notebook examples} containing a conversational prompt format used in the context of other QA tasks. We therefore use this format by default but search over the prompt formats for model-specific prompt selection, since it is not specifically designed for closed-ended multiple-choice QA tasks. Moreover, as the system prompt provided is not semantically applicable to the QA tasks that we consider (e.g., \texttt{``You are working as an excellent assistant in chemistry and molecule discovery.''}, we use our own manually designed default system prompt. We include the default prompt format used for \textsc{BioMedGPT-LM} in Appendix~\ref{sec:biomedgpt-template}.

\paragraph{\textsc{LLaVA-Med} \citep{llava-med}.} For \textsc{LLaVA-Med}, we use the system prompt and conversational prompt format included in the \href{https://github.com/microsoft/LLaVA-Med/blob/b9a98a736d2ef05bcf5ff345be6403fb3a664eaf/llava/conversation.py#L243}{``simple\_conv\_med'' template} from the official GitHub repository (for \textsc{LLaVA-v0} \citep{llava}, we use the \href{https://github.com/microsoft/LLaVA-Med/blob/b9a98a736d2ef05bcf5ff345be6403fb3a664eaf/llava/conversation.py#L257}{``simple\_conv'' template}) by default.
For formatting the visual questions, we also refer to \href{https://github.com/microsoft/LLaVA-Med/blob/b9a98a736d2ef05bcf5ff345be6403fb3a664eaf/llava/eval/eval_metrics/answer-file-llava-zeorshot.jsonl}{this file} containing the raw visual QA results on VQA-RAD (\texttt{``Please choose from the following two options: [yes, no]''}).
Meanwhile, we make these choices with the following caveats, to the best of our knowledge. 
First, the exact choice of system prompt and conversational prompt format used for evaluation are not discussed in the paper or the code repository, and we choose the one that has a system prompt specific to \textsc{LLaVA-Med} (\texttt{``You are LLaVA-Med, a large language and vision assistant trained by a group of researchers at Microsoft \ldots''}) and follows the conversational format used for \textsc{Vicuna-v0} \citep{vicuna}, which forms its LLM backbone. Second, details on how the answer choices should be formatted in the context of closed-ended QA tasks is only shown in the VQA-RAD results file.
Given the uncertainty in such details, we also search over the prompt formats for model-specific prompt selection.
We note that \textsc{LLaVA-Med} was not pretrained on multi-image inputs or evaluated in few-shot setting, and therefore details on the choice of few-shot examples are irrelevant.
For sampling, the \href{https://github.com/microsoft/LLaVA-Med/blob/b9a98a736d2ef05bcf5ff345be6403fb3a664eaf/llava/eval/model_vqa_med.py}{evaluation code} uses a default temperature setting of 0.7 (albeit without the random seeds), which differs from our evaluation setup. We include the default prompt format used for \textsc{LLaVA-Med} in Appendix~\ref{sec:llava-med-template} and that for \textsc{LLaVA-v0} in Appendix \ref{sec:llava-template}.

\paragraph{\textsc{Med-Flamingo} \citep{med-flamingo}.} For \textsc{Med-Flamingo}, we use the system prompt and prompt format provided in the \href{https://github.com/snap-stanford/med-flamingo/blob/7bcbb6c3932c814a6f7ae4b838ae4fada39b42a4/scripts/demo.py}{demo code} from the official GitHub repository by default.
However, we search over the prompt formats when performing model-specific prompt selection, as the example prompt in the demo does not show details for formatting answer choices in a closed-ended QA context.
The choice of few-shot examples and the sampling details used for the original evaluations on VQA-RAD and PathVQA are not available.
We include the default prompt format used for \textsc{Med-Flamingo} in Appendix~\ref{sec:med-flamingo-template} and that for \textsc{Open-Flamingo} in Appendix~\ref{sec:open-flamingo-template}.

\subsection{Default LLM Prompt Formats}\label{sec:llm-prompt-templates}
In this section, we share the \textit{default} prompt formats that we use for each LLM, using MMLU (Clinical Knowledge) \citep{mmlu} as a running example. We denote the system prompt in \system{red}, any few-shot examples in \example{green}, and the question being asked of the model in \question{purple}.

For models that do not have a specific system prompt and prompt format designed for closed-ended medical QA (see Section \ref{sec:reproducibility}), we use a manually designed prompt format by default. This includes all of the general-domain LLMs. For example, in the 1-shot setting, the default prompt for non-instruction-tuned models is as follows:

\begin{mdframed}
\system{The following is a multiple-choice question about medical knowledge. Answer the question by choosing one of the options from A to D.}

\noindent \example{\#\#\# Question: Glycolysis is the name given to the pathway involving the conversion of:\\ 
(A) glycogen to glucose-1-phosphate.\\
(B) glycogen or glucose to fructose.\\
(C) glycogen or glucose to pyruvate or lactate.\\
(D) glycogen or glucose to pyruvate or acetyl CoA.\\
\noindent \#\#\# Answer: (C) glycogen or glucose to pyruvate or lactate.}

\noindent \question{\#\#\# Question: What size of cannula would you use in a patient who needed a rapid blood transfusion (as of 2020 medical knowledge)?\\
(A) 18 gauge.\\
(B) 20 gauge.\\
(C) 22 gauge.\\
(D) 24 gauge.\\
\noindent \#\#\# Answer:}
\end{mdframed}

\noindent For instruction-tuned models, which typically expect a specific \textit{conversational} format, we apply the above format to each ``user'' query and ``assistant'' response and remove the \texttt{`\#\#\#'} and \texttt{`Answer:'} tags. For example, the input prompt to \textsc{Llama-3-70B-Instruct} is as follows:

\begin{mdframed}
<|begin\_of\_text|> \\
<|start\_header\_id|> system <|end\_header\_id|> \\
\system{The following is a multiple-choice question about medical knowledge. Answer the question by choosing one of the options from A to D.}<|eot\_id|>\\
<|start\_header\_id|>user<|end\_header\_id|>
\noindent \example{Question: Glycolysis is the name given to the pathway involving the conversion of:\\ 
(A) glycogen to glucose-1-phosphate.\\
(B) glycogen or glucose to fructose.\\
(C) glycogen or glucose to pyruvate or lactate.\\
(D) glycogen or glucose to pyruvate or acetyl CoA.}<|eot\_id|>\\
<|start\_header\_id|>assistant<|end\_header\_id|>\\
\noindent \example{(C) glycogen or glucose to pyruvate or lactate.}<|eot\_id|>\\
<|start\_header\_id|>user<|end\_header\_id|>

\noindent \question{Question: What size of cannula would you use in a patient who needed a rapid blood transfusion (as of 2020 medical knowledge)?\\
(A) 18 gauge.\\
(B) 20 gauge.\\
(C) 22 gauge.\\
(D) 24 gauge.}<|eot\_id|>\\
<|start\_header\_id|>assistant<|end\_header\_id|>
\end{mdframed}
In the following subsections, we show the system prompt and prompt formats used in the 1-shot setting for models that have a dedicated format. We exclude the model-specific special tokens (e.g., \texttt{`[INST]'}) for ease of presentation, and add \texttt{`[User]'} and \texttt{`[Model]'} to demarcate each question and answer for the instruction-tuned models.

\subsubsection{\textsc{OpenBioLLM} \citep{OpenBioLLMs}}\label{sec:openbiollm-template}
\begin{mdframed}
\system{You are an expert and experienced from the healthcare and biomedical domain with extensive medical knowledge and practical experience. Your name is OpenBioLLM, and you were developed by Saama AI Labs. who's willing to help answer the user's query with explanation. In your explanation, leverage your deep medical expertise such as relevant anatomical structures, physiological processes, diagnostic criteria, treatment guidelines, or other pertinent medical concepts. Use precise medical terminology while still aiming to make the explanation clear and accessible to a general audience.}

\noindent \textbf{[User]} \example{Question: Glycolysis is the name given to the pathway involving the conversion of:\\
(A) glycogen to glucose-1-phosphate.\\
(B) glycogen or glucose to fructose.\\
(C) glycogen or glucose to pyruvate or lactate.\\
(D) glycogen or glucose to pyruvate or acetyl CoA.}\\
\example{
\textcolor{black}{\textbf{[Model]}} (C) glycogen or glucose to pyruvate or lactate.
}

\noindent \textbf{[User]} \question{Question: What size of cannula would you use in a patient who needed a rapid blood transfusion (as of 2020 medical knowledge)?\\
(A) 18 gauge.\\
(B) 20 gauge.\\
(C) 22 gauge.\\
(D) 24 gauge.
}

\end{mdframed}

\subsubsection{\textsc{Clinical-Camel} \citep{clinical-camel}}\label{sec:clinical-camel-template}
\begin{mdframed}
\system{The following is a multiple-choice question about medical knowledge. Answer the question by choosing one of the options from A to D.}\\
\example{\textcolor{black}{\textbf{[User]}} Question: Glycolysis is the name given to the pathway involving the conversion of:\\
(A) glycogen to glucose-1-phosphate.\\
(B) glycogen or glucose to fructose.\\
(C) glycogen or glucose to pyruvate or lactate.\\
(D) glycogen or glucose to pyruvate or acetyl CoA.
}\\
\example{\textcolor{black}{\textbf{[Model]}} (C) glycogen or glucose to pyruvate or lactate.}\\ \question{\textcolor{black}{\textbf{[User]}} Question: What size of cannula would you use in a patient who needed a rapid blood transfusion (as of 2020 medical knowledge)?\\
(A) 18 gauge.\\
(B) 20 gauge.\\
(C) 22 gauge.\\
(D) 24 gauge.
}
\end{mdframed}

\subsubsection{\textsc{BioMistral} \citep{biomistral}}\label{sec:biomistral-template}
\begin{mdframed}
\system{The following are multiple choice questions (with answers) about medical knowledge.}

\noindent \example{**Question:** Glycolysis is the name given to the pathway involving the conversion of:\\
(A) glycogen to glucose-1-phosphate.\\
(B) glycogen or glucose to fructose.\\
(C) glycogen or glucose to pyruvate or lactate.\\
(D) glycogen or glucose to pyruvate or acetyl CoA.\\
**Answer:** (C
}

\noindent \question{**Question:** What size of cannula would you use in a patient who needed a rapid blood transfusion (as of 2020 medical knowledge)?\\
(A) 18 gauge.\\
(B) 20 gauge.\\
(C) 22 gauge.\\
(D) 24 gauge.\\
**Answer:** (
}
\end{mdframed}

\subsubsection{\textsc{MediTron} \citep{meditron}}\label{sec:meditron-template}
\begin{mdframed}
\system{You are a medical doctor answering real-world medical entrance exam questions. Based on your understanding of basic and clinical science, medical knowledge, and mechanisms underlying health, disease, patient care, and modes of therapy, answer the following multiple-choice question. Select one correct answer from A to D. Base your answer on the current and standard practices referenced in medical guidelines.}

\noindent \example{Question: Glycolysis is the name given to the pathway involving the conversion of:\\
Options:\\
A. glycogen to glucose-1-phosphate.\\
B. glycogen or glucose to fructose.\\
C. glycogen or glucose to pyruvate or lactate.\\
D. glycogen or glucose to pyruvate or acetyl CoA.\\
The answer is: C}\\
\noindent \question{Question: What size of cannula would you use in a patient who needed a rapid blood transfusion (as of 2020 medical knowledge)?\\
Options:\\
A. 18 gauge.\\
B. 20 gauge.\\
C. 22 gauge.\\
D. 24 gauge.\\
The answer is:
}
\end{mdframed}

\subsubsection{\textsc{BioMedGPT-LM} \citep{biomedgpt}}\label{sec:biomedgpt-template}
\begin{mdframed}
\system{The following is a multiple-choice question about medical knowledge. Answer the question by choosing one of the options from A to D.}

\noindent \example{\#\#\# Human: Glycolysis is the name given to the pathway involving the conversion of:\\
(A) glycogen to glucose-1-phosphate.\\
(B) glycogen or glucose to fructose.\\
(C) glycogen or glucose to pyruvate or lactate.\\
(D) glycogen or glucose to pyruvate or acetyl CoA.\\
\#\#\# Assistant: (C) glycogen or glucose to pyruvate or lactate.
}

\noindent \question{\#\#\# Human: What size of cannula would you use in a patient who needed a rapid blood transfusion (as of 2020 medical knowledge)?\\
(A) 18 gauge.\\
(B) 20 gauge.\\
(C) 22 gauge.\\
(D) 24 gauge.\\
\#\#\# Assistant: 
}
\end{mdframed}

\subsection{Default VLM Prompt Formats}\label{sec:vlm-prompt-templates}
In this section, we share the \textit{default} prompt formats that we use for each general-domain/medical VLM, using VQA-RAD \citep{vqa-rad} as a running example. We denote the system prompt in \system{red}, any few-shot examples in \example{green}, and the question being asked of the model in \question{purple}.
By default, we show the format used in the 1-shot setting.

\subsubsection{\textsc{LLaVA-Med} \citep{llava-med}}\label{sec:llava-med-template}

\begin{mdframed}
\system{You are LLaVA-Med, a large language and vision assistant trained by a group of researchers at Microsoft, based on the general domain LLaVA architecture. You are able to understand the visual content that the user provides, and assist the user with a variety of medical and clinical tasks using natural language.}\\
\system{Follow the instructions carefully and explain your answers in detail.}

\noindent \example{\#\#\# Human: Does this patient have multiple lesions in their chest? Please choose from the following options: [yes, no]. <image>\\
\#\#\# Assistant: no
}

\noindent \question{\#\#\# Human: Is there evidence of an aortic aneurysm? Please choose from the following options: [yes, no]. <image>\\
\#\#\# Assistant: 
}
\end{mdframed}

\subsubsection{\textsc{LLaVA-v0} \citep{llava}}\label{sec:llava-template}
\begin{mdframed}
\system{A chat between a curious human and an artificial intelligence assistant. The assistant gives helpful, detailed, and polite answers to the human's questions.}

\noindent \example{\#\#\# Human: Does this patient have multiple lesions in their chest? Please choose from the following options: [yes, no]. <image>\\
\#\#\# Assistant: no
}

\noindent \question{\#\#\# Human: Is there evidence of an aortic aneurysm? Please choose from the following options: [yes, no]. <image>\\
\#\#\# Assistant:
}
\end{mdframed}

\subsubsection{\textsc{Med-Flamingo} \citep{med-flamingo}}\label{sec:med-flamingo-template}
\begin{mdframed}
\system{You are a helpful medical assistant. You are being provided with images, a question about the image and an answer. Follow the examples and answer the last question.}

\noindent \example{<image> Does this patient have multiple lesions in their chest?\\
(A) yes\\
(B) no\\
Answer: (B) no <|endofchunk|>
}

\noindent \question{<image> Is there evidence of an aortic aneurysm?\\
(A) yes\\
(B) no\\
Answer: 
}
\end{mdframed}

\subsubsection{\textsc{Open-Flamingo} \citep{open-flamingo}}\label{sec:open-flamingo-template}
\begin{mdframed}
\system{The following is a multiple-choice visual question requiring medical knowledge. Answer the question by choosing one of the provided answer options.}

\noindent \example{<image> Does this patient have multiple lesions in their chest?\\
(A) yes\\
(B) no}\\
\example{Answer: (B) no <|endofchunk|>
}
\noindent \question{<image> Is there evidence of an aortic aneurysm?\\
(A) yes\\
(B) no\\
Answer: 
}
\end{mdframed}

%% file: latex/sections/appendix_results.tex
\begin{figure*}[t!]
    \centering
    \begin{tabular}{@{}c@{}c@{}}
        \multicolumn{2}{c}{
            \begin{subfigure}{0.98\linewidth}
                \includegraphics[width=\linewidth]{figs/llm-acc-ci-legend.pdf}
            \end{subfigure}
        }
        \\
        \begin{subfigure}{0\linewidth}
        \end{subfigure} &
        \begin{subfigure}{0.98\linewidth}
            \includegraphics[width=\linewidth]{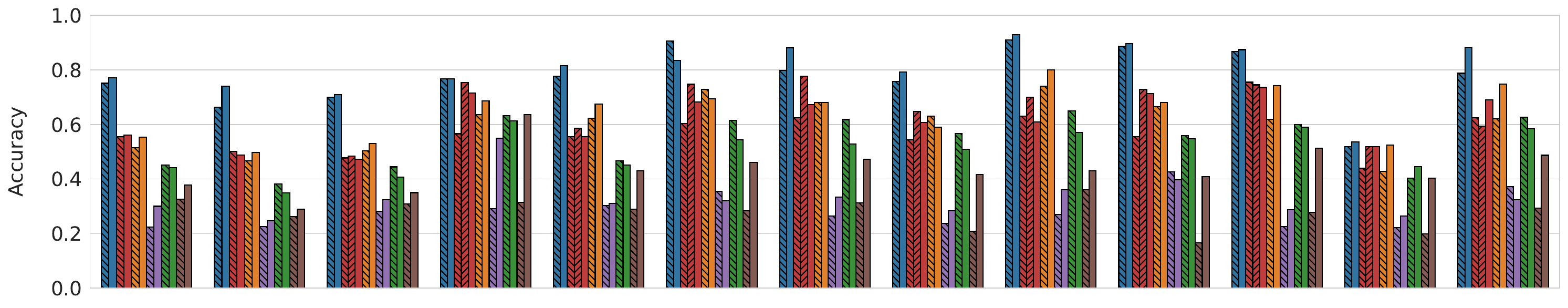}
        \end{subfigure}
        \\
        \begin{subfigure}{0\linewidth}
        \end{subfigure} &
        \begin{subfigure}{0.98\linewidth}
            \includegraphics[width=\linewidth]{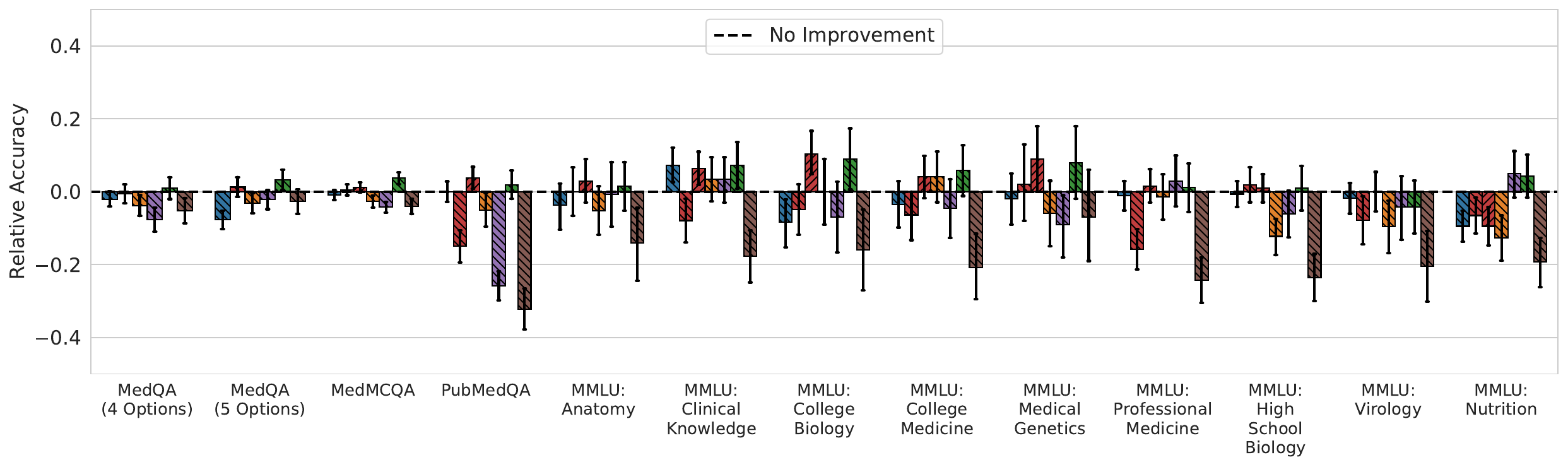}
        \end{subfigure}
    \end{tabular}
    \caption{Medical LLMs do not show a statistically significant improvement over their general-domain counterparts in the zero-shot setting, after independently selecting the best prompt format and examples for each model. Top row shows the absolute exact-match accuracies on the test set, and bottom row shows the relative exact-match accuracies along with 95\% confidence intervals derived via bootstrapping on the test set (see Section~\ref{sec:eval-setup}). We show the results for when model predictions are generated via greedy decoding.}
    \label{fig:llm-0-acc-ci}
\end{figure*}

\begin{figure*}[t!]
    \centering
    \begin{tabular}{@{}c@{}c@{}}
        \multicolumn{2}{c}{
            \begin{subfigure}{0.98\linewidth}
                \includegraphics[width=\linewidth]{figs/llm-acc-ci-legend.pdf}
            \end{subfigure}
        }
        \\
        \begin{subfigure}{0.02\linewidth}
            \makebox[\linewidth]{\raisebox{0pt}{{(a)}}}
        \end{subfigure} &
        \begin{subfigure}{0.96\linewidth}
            \includegraphics[width=\linewidth]{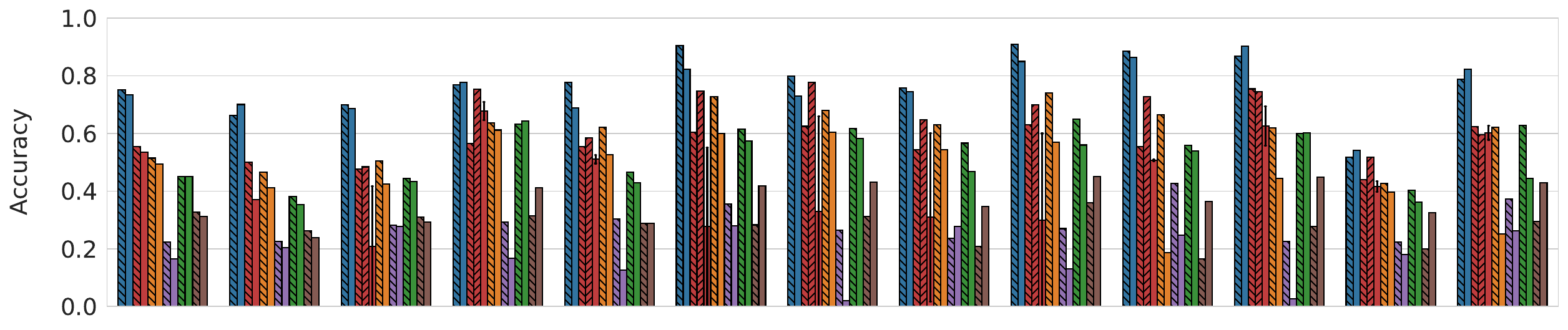}
        \end{subfigure}
        \\
        \begin{subfigure}{0.02\linewidth}
        \end{subfigure} &
        \begin{subfigure}{0.96\linewidth}
            \includegraphics[width=\linewidth]{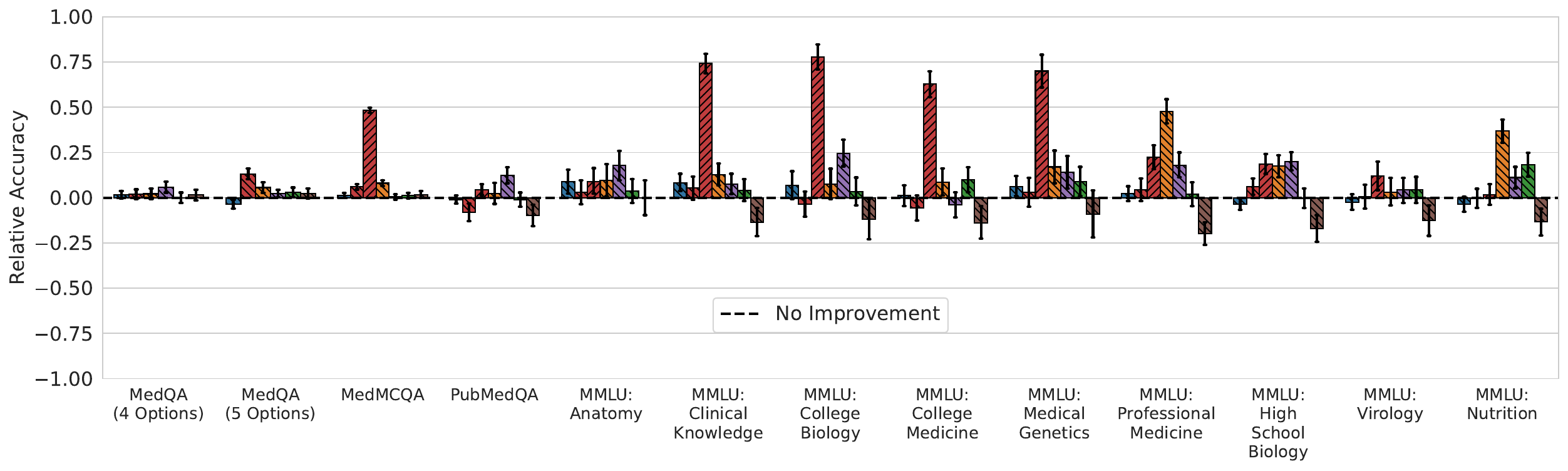}
        \end{subfigure}
        \\
        \begin{subfigure}{0.02\linewidth}
            \makebox[\linewidth]{\raisebox{0pt}{{(b)}}}
        \end{subfigure} &
        \begin{subfigure}{0.96\linewidth}
            \includegraphics[width=\linewidth]{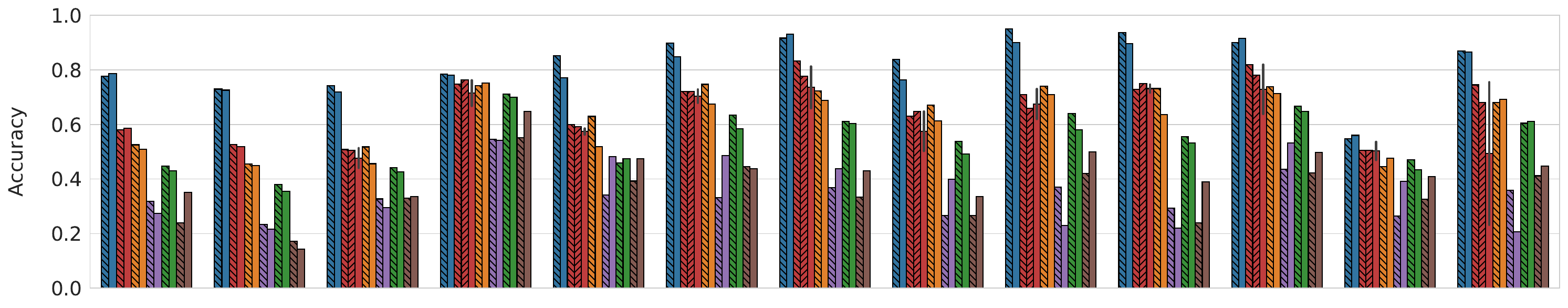}
        \end{subfigure}
        \\
        \begin{subfigure}{0.02\linewidth}
        \end{subfigure} &
        \begin{subfigure}{0.96\linewidth}
            \includegraphics[width=\linewidth]{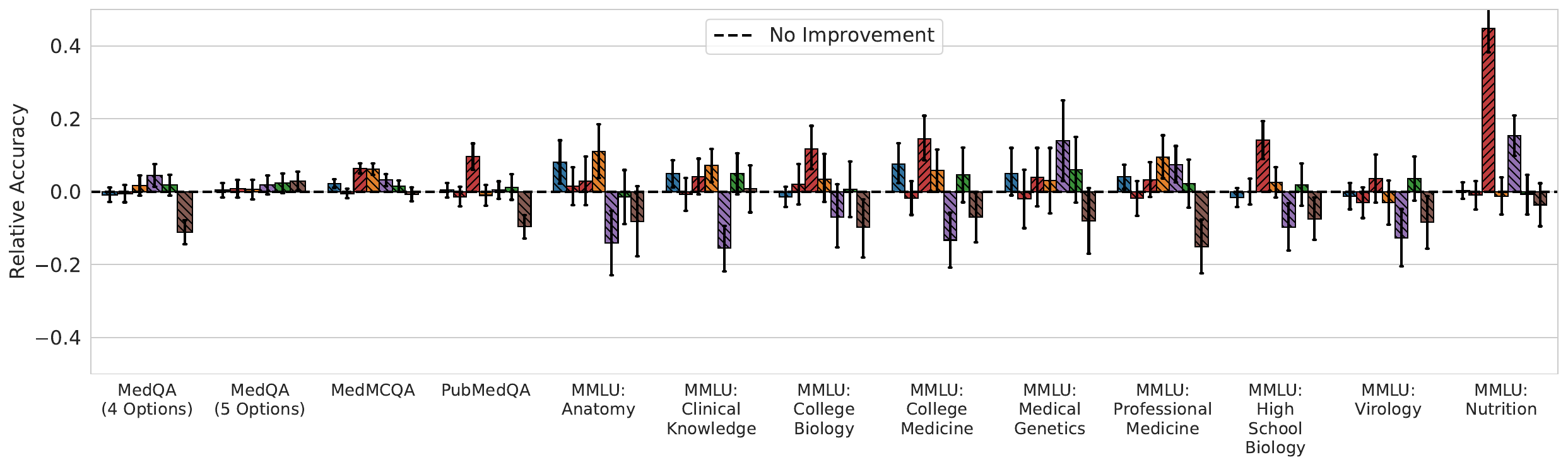}
        \end{subfigure}
    \end{tabular}
    \caption{Using a single, fixed prompt format only optimized for the medical model can overestimate the performance improvements from medical DAPT, in both (a) zero-shot and (b) 3-shot settings. For each setting, top row shows the absolute exact-match accuracies on the test set, and bottom row shows the relative exact-match accuracies along with 95\% confidence intervals derived via bootstrapping on the test set (see Section~\ref{sec:eval-setup}). For \textsc{Llama-2-70B}, which has multiple corresponding medical LLMs (\textsc{MediTron-70B} and \textsc{Clinical-Camel-70B}), we include a min-max error bar in the absolute accuracy plots to show how the absolute accuracy changes with respect to each prompt.
    We show the results for when model predictions are generated via greedy decoding.}
    \label{fig:llm-acc-ci-med}
\end{figure*}

\begin{figure*}[t!]
    \centering
    \begin{tabular}{@{}c@{}c@{\hskip 2pt}c@{}c@{}}
        \multicolumn{4}{c}{
            \begin{subfigure}{0.7\linewidth}
                \includegraphics[width=\linewidth]{figs/vlm-acc-ci-legend-v2.pdf}
            \end{subfigure}
        }
        \\
        \begin{subfigure}{0.03\linewidth}
            \makebox[\linewidth]{\raisebox{60pt}{{(a)}}}
        \end{subfigure} &
        \begin{subfigure}{0.47\linewidth}
            \includegraphics[width=\linewidth]{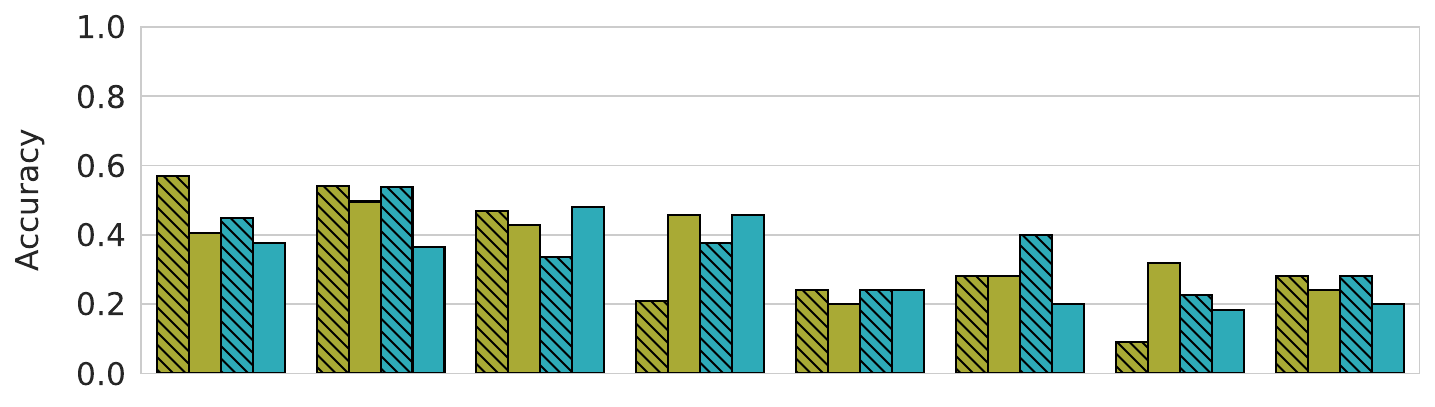}
        \end{subfigure} &
        \begin{subfigure}{0.03\linewidth}
            \makebox[\linewidth]{\raisebox{60pt}{{(b)}}}
        \end{subfigure} &
        \begin{subfigure}{0.47\linewidth}
            \includegraphics[width=\linewidth]{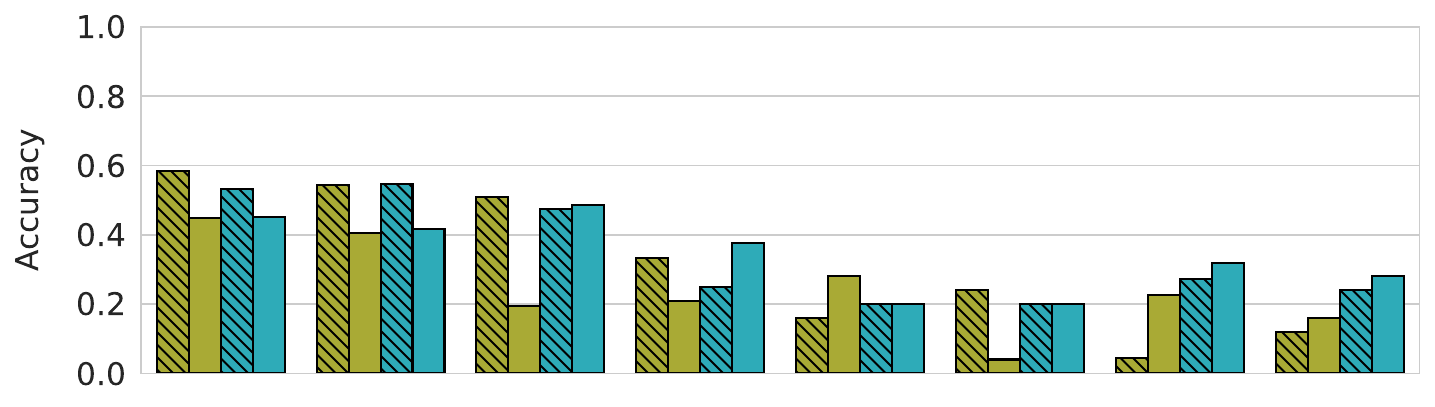}
        \end{subfigure}
        \\
        \begin{subfigure}{0.03\linewidth}
        \end{subfigure} &
        \begin{subfigure}{0.47\linewidth}
            \includegraphics[width=\linewidth]{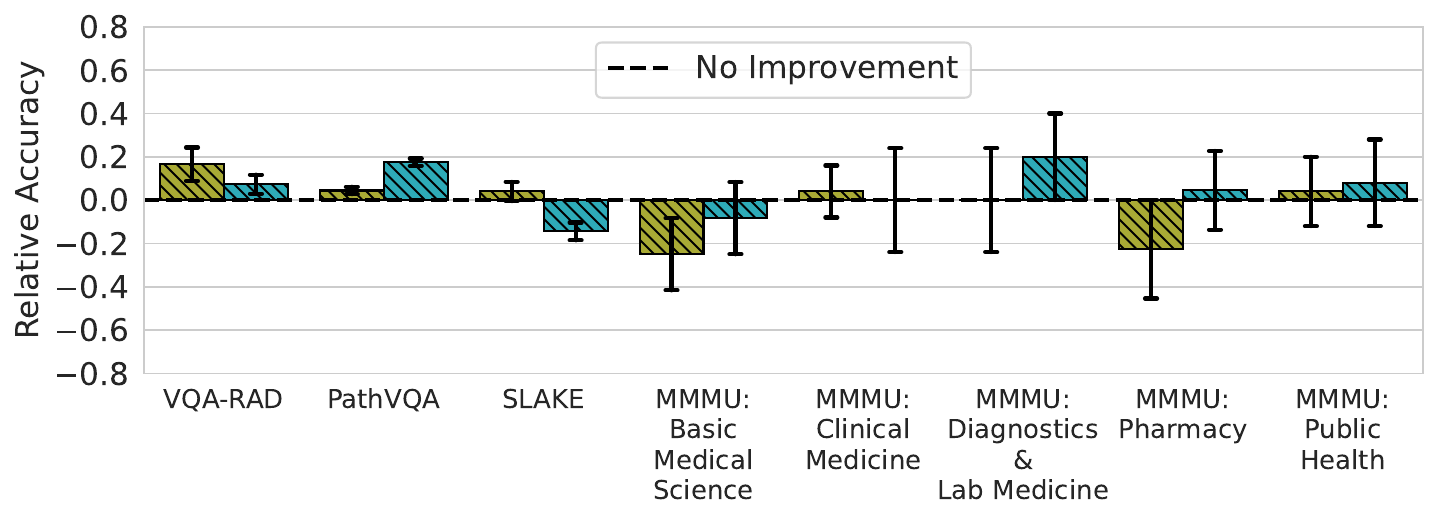}
        \end{subfigure} &
        \begin{subfigure}{0.03\linewidth}
        \end{subfigure} &
        \begin{subfigure}{0.47\linewidth}
            \includegraphics[width=\linewidth]{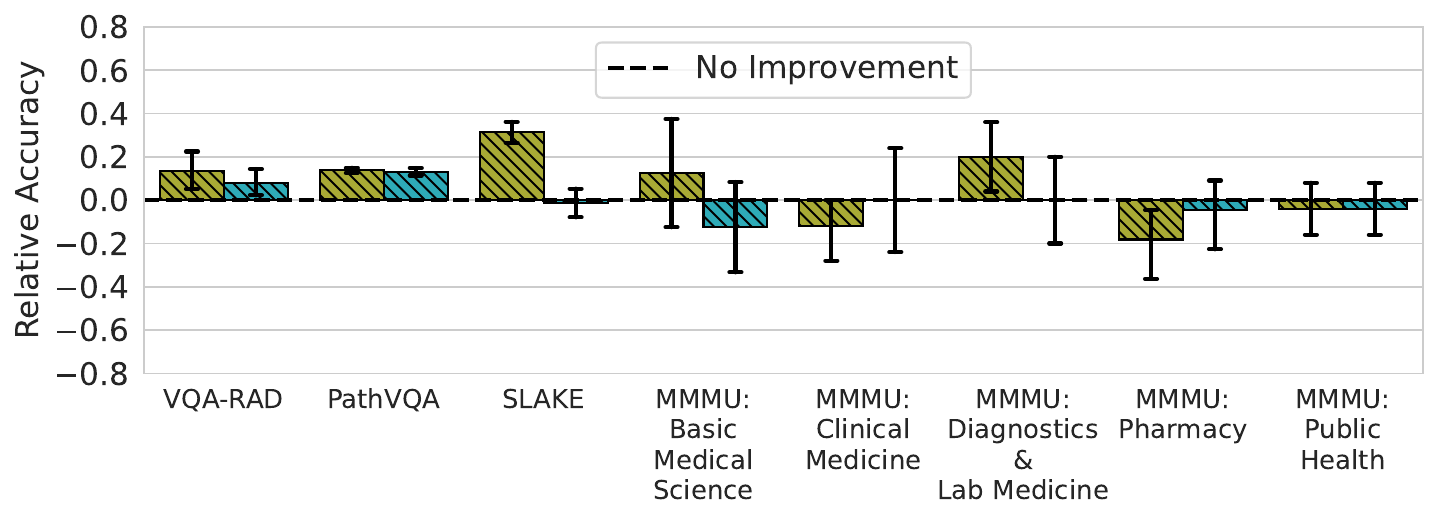}
        \end{subfigure}
    \end{tabular}
    \vspace{-10pt}
    \caption{Using a single, fixed prompt format only optimized for the medical model can overestimate the performance improvements from medical DAPT, in both (a) zero-shot and (b) 3-shot settings. For each setting, top row shows the raw exact-match accuracies on the test set, and the bottom row shows the \textit{relative} exact-match accuracies along with 95\% confidence intervals derived via boostrapping on the test set (see Section~\ref{sec:eval-setup}). We show the results for when model predictions are generated via greedy decoding.}
    \label{fig:vlm-acc-ci-med}
\end{figure*}

In this section, we provide additional results for the main zero-/few-shot prompting experiments with greedy decoding, which are discussed in Section \ref{sec:results}. 

\subsection{Finding 1 (Section~\ref{sec:results})}
\label{sec:greedy-decoding-finding1}

\begin{table}[t!]
    \centering
    \caption{The win, tie, and loss rates (\%) of all medical LLMs (top) and VLMs (bottom) in the zero-shot setting, after independently optimizing the prompts for both medical and general-domain models. Model predictions are generated via greedy decoding.}
    \label{tab:win-tie-loss-rates-0}
    \resizebox{\linewidth}{!}{
    
    \begin{tabular}{@{}l@{\hskip 7pt}c@{\hskip 7pt}c@{\hskip 7pt}c@{\hskip 3pt}}
        \toprule
        Model & Win & Tie & Loss \\
        \midrule
        \textsc{OpenBioLLM-70B} \citep{OpenBioLLMs} & 7.7 & 69.2 & \textbf{23.1} \\
        \textsc{MediTron-70B} \citep{meditron} & 0 & 61.5 & \textbf{38.5} \\
        \textsc{Clinical-Camel-70B} \citep{clinical-camel} & \textbf{27.3} & 63.6 & 9.1 \\
        \textsc{OpenBioLLM-8B} \citep{OpenBioLLMs} & 0 & 46.2 & \textbf{53.8} \\
        \textsc{MediTron-7B} \citep{meditron} & 0 & 69.2 & \textbf{30.8} \\
        \textsc{BioMistral-7B} \citep{biomistral} & \textbf{30.8} & 69.2 & 0 \\
        \textsc{BioMedGPT-LM-7B} \citep{biomedgpt} & 0 & 15.4 & \textbf{84.6} \\
        \midrule
        \textsc{LLaVA-Med-7B} \citep{llava-med} & 12.5 & 62.5 & \textbf{25.0} \\
        \textsc{Med-Flamingo-9B} \citep{med-flamingo} & 0 & 87.5 & \textbf{12.5} \\
        \bottomrule
    \end{tabular}
    }
\end{table}

\begin{table}[t!]
    \centering
    \caption{The win, tie, and loss rates (\%) of all medical LLMs (top) and VLMs (bottom) in the 3-shot setting, after independently optimizing the prompts for both medical and general-domain models. Model predictions are generated via greedy decoding.}
    \label{tab:win-tie-loss-rates-3}
    \resizebox{\linewidth}{!}{
    
    \begin{tabular}{@{}l@{\hskip 7pt}c@{\hskip 7pt}c@{\hskip 7pt}c@{\hskip 3pt}}
        \toprule
        Model & Win & Tie & Loss \\
        \midrule
        \textsc{OpenBioLLM-70B} \citep{OpenBioLLMs} & \textbf{30.8} & 69.2 & 0 \\
        \textsc{MediTron-70B} \citep{meditron} & 0 & 69.2 & \textbf{30.8} \\
        \textsc{Clinical-Camel-70B} \citep{clinical-camel} & 0 & 63.6 & \textbf{36.4} \\
        \textsc{OpenBioLLM-8B} \citep{OpenBioLLMs} & 7.7 & 61.5 & \textbf{30.8} \\
        \textsc{MediTron-7B} \citep{meditron} & 0 & 23.1 & \textbf{76.9} \\
        \textsc{BioMistral-7B} \citep{biomistral} & \textbf{46.2} & 53.8 & 0 \\
        \textsc{BioMedGPT-LM-7B} \citep{biomedgpt} & 0 & 7.7 & \textbf{92.3} \\
        \midrule
        \textsc{LLaVA-Med-7B} \citep{llava-med} & 12.5 & 62.5 & \textbf{25.0} \\
        \textsc{Med-Flamingo-9B} \citep{med-flamingo} & 0 & 100.0 & 0 \\
        \bottomrule
    \end{tabular}
    }
\end{table}

Figure~\ref{fig:llm-0-acc-ci} shows the absolute and relative exact-match accuracies achieved by the medical and general-domain LLMs in the zero-shot prompting regime, after independently optimizing the prompt for each model. In Tables~\ref{tab:win-tie-loss-rates-0}--\ref{tab:win-tie-loss-rates-3}, we also show the zero-shot and 3-shot win/tie/loss rates achieved by the medical LLMs and VLMs. 
For \textsc{Clinical-Camel-70B}, we compute the win/tie/loss rates while excluding the MedQA datasets, as discussed in Section~\ref{sec:results}.  
For each medical model, we boldface the win rate if it wins more than it loses to its general-domain base model, and vice versa. 

As discussed in Finding 1 of Section~\ref{sec:results}, we find that in both the zero-shot and 3-shot settings, only 2 out of 7 medical models show statistically significant improvements over their corresponding base models (\textsc{Clinical-Camel-70B} and \textsc{BioMistral-7B} for zero-shot; \textsc{OpenBioLLM-70B} and \textsc{BioMistral-7B} for 3-shot), albeit by a limited margin in terms of absolute accuracy. 
For all other models, the win rates are less than or equal to the loss rates, and the majority of cases result in a tie (i.e., the confidence interval crosses zero relative accuracy). 

\subsection{Finding 2 (Section~\ref{sec:results})}
\label{sec:greedy-decoding-finding2}

\begin{table}[t!]
    \centering
    \caption{The win, tie, and loss rates (\%) of all medical LLMs (top) and VLMs (bottom) in the zero-shot setting, when using a single, fixed prompt optimized only for the medical model. Model predictions are generated via greedy decoding.}
    \label{tab:win-tie-loss-rates-med-0}
    \resizebox{\linewidth}{!}{
    
    \begin{tabular}{@{}l@{\hskip 7pt}c@{\hskip 7pt}c@{\hskip 7pt}c@{\hskip 3pt}}
        \toprule
        Model & Win & Tie & Loss \\
        \midrule
        \textsc{OpenBioLLM-70B} \citep{OpenBioLLMs} & \textbf{23.1} & 61.5 & 15.4 \\
        \textsc{MediTron-70B} \citep{meditron} & \textbf{23.1} & 69.2 & 7.7 \\
        \textsc{OpenBioLLM-8B} \citep{OpenBioLLMs} & \textbf{69.2} & 30.8 & 0 \\
        \textsc{MediTron-7B} \citep{meditron} & \textbf{76.9} & 23.1 & 0 \\
        \textsc{BioMistral-7B} \citep{biomistral} & \textbf{30.8} & 69.2 & 0 \\
        \textsc{BioMedGPT-LM-7B} \citep{biomedgpt} & 0 & 38.5 & \textbf{61.5} \\
        \midrule
        \textsc{LLaVA-Med-7B} \citep{llava-med} & \textbf{25.0} & 62.5 & 12.5 \\
        \textsc{Med-Flamingo-9B} \citep{med-flamingo} & \textbf{25.0} & 62.5 & 12.5 \\
        \bottomrule
    \end{tabular}
    }
\end{table}

\begin{table}[t!]
    \centering
    \caption{The win, tie, and loss rates (\%) of all medical LLMs (top) and VLMs (bottom) in the 3-shot setting, when using a single, fixed prompt optimized only for the medical model. Model predictions are generated via greedy decoding.}
    \label{tab:win-tie-loss-rates-med-3}
    \resizebox{\linewidth}{!}{
    
    \begin{tabular}{@{}l@{\hskip 7pt}c@{\hskip 7pt}c@{\hskip 7pt}c@{\hskip 3pt}}
        \toprule
        Model & Win & Tie & Loss \\
        \midrule
        \textsc{OpenBioLLM-70B} \citep{OpenBioLLMs} & \textbf{38.5} & 61.5 & 0 \\
        \textsc{MediTron-70B} \citep{meditron} & 0 & 100.0 & 0 \\
        \textsc{Clinical-Camel-70B} \citep{clinical-camel} & \textbf{54.5} & 45.5 & 0 \\
        \textsc{OpenBioLLM-8B} \citep{OpenBioLLMs} & \textbf{30.8} & 69.2 & 0 \\
        \textsc{MediTron-7B} \citep{meditron} & 38.5 & 23.1 & 38.5 \\
        \textsc{BioMistral-7B} \citep{biomistral} & 0 & 100.0 & 0 \\
        \textsc{BioMedGPT-LM-7B} \citep{biomedgpt} & 7.7 & 46.2 & \textbf{46.2} \\
        \midrule
        \textsc{LLaVA-Med-7B} \citep{llava-med} & \textbf{50.0} & 37.5 & 12.5 \\
        \textsc{Med-Flamingo-9B} \citep{med-flamingo} & \textbf{25.0} & 75.0 & 0 \\
        \bottomrule
    \end{tabular}
    }
\end{table}

Figures~\ref{fig:llm-acc-ci-med}--\ref{fig:vlm-acc-ci-med} show how the absolute and relative exact-match accuracies change for LLMs and VLMs in the zero-shot and 3-shot settings, when we use a single, fixed prompt that is only optimized for the medical model. 
In Tables~\ref{tab:win-tie-loss-rates-med-0}--\ref{tab:win-tie-loss-rates-med-3}, we also show the zero-shot and 3-shot win/tie/loss rates in this scenario. 
For each medical model, we boldface the win rate if it wins more than it loses to its general-domain base model, and vice versa. 

Compared to when the prompt is independently optimized for each model, we see that a greater number of medical models show statistically significant improvements. 
Notably, all medical VLMs outperform their general-domain counterparts in both zero-shot and few-shot accuracy under this setup, and all but one medical LLMs outperform their general-domain counterparts in the zero-shot setting.
These results suggest that using a single, fixed prompt that is only tailored to one model can result in an unfair comparison and can potentially lead to an overestimation of the performance benefits of medical DAPT.

%% file: latex/sections/appendix_constrained_results.tex
In this section, we provide all of the results for the zero-/few-shot prompting evaluations described in Section \ref{sec:eval-setup}, where the model predictions are generated via \textit{constrained} instead of greedy decoding.

\subsection{Finding 1 (Section~\ref{sec:results})}
\label{sec:constrained-decoding-finding1}

\begin{figure*}[t!]
    \centering
    \begin{tabular}{@{}c@{}c@{}}
        \multicolumn{2}{c}{
            \begin{subfigure}{0.98\linewidth}
                \includegraphics[width=\linewidth]{figs/llm-acc-ci-legend.pdf}
            \end{subfigure}
        }
        \\
        \begin{subfigure}{0.02\linewidth}
            \makebox[\linewidth]{\raisebox{0pt}{{(a)}}}
        \end{subfigure} &
        \begin{subfigure}{0.96\linewidth}
            \includegraphics[width=\linewidth]{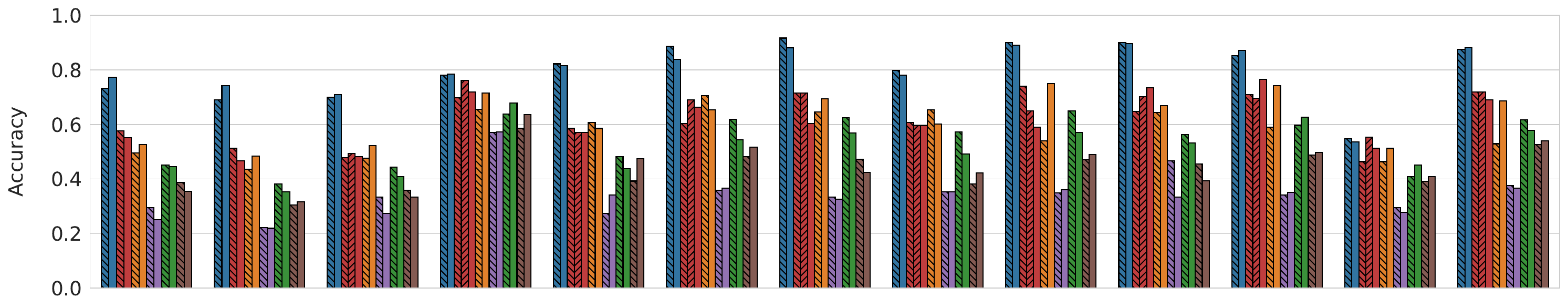}
        \end{subfigure}
        \\
        \begin{subfigure}{0.02\linewidth}
        \end{subfigure} &
        \begin{subfigure}{0.96\linewidth}
            \includegraphics[width=\linewidth]{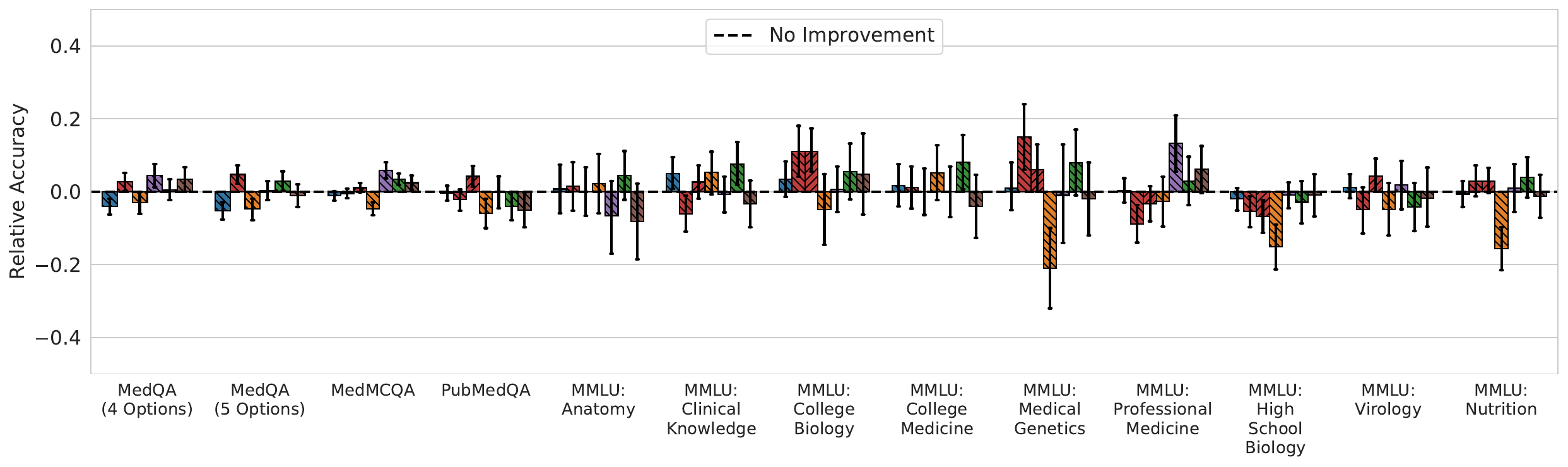}
        \end{subfigure}
        \\
        \begin{subfigure}{0.02\linewidth}
            \makebox[\linewidth]{\raisebox{0pt}{{(b)}}}
        \end{subfigure} &
        \begin{subfigure}{0.96\linewidth}
            \includegraphics[width=\linewidth]{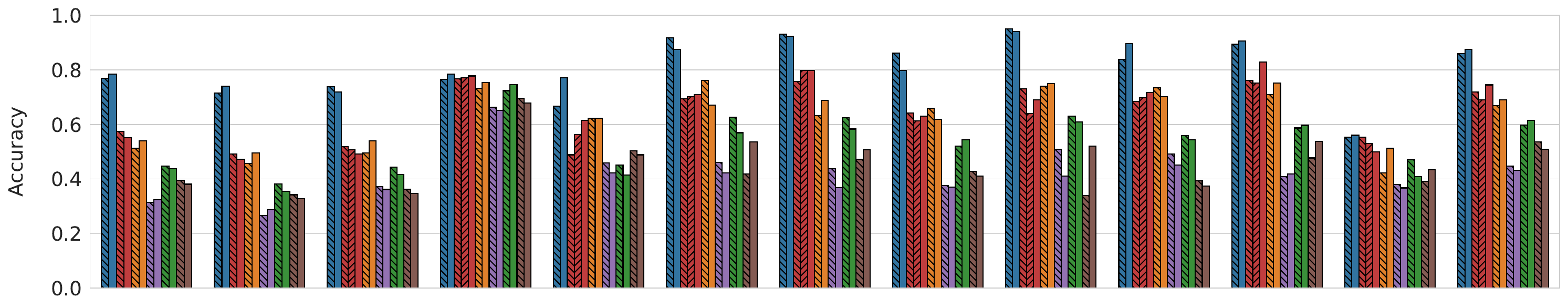}
        \end{subfigure}
        \\
        \begin{subfigure}{0.02\linewidth}
        \end{subfigure} &
        \begin{subfigure}{0.96\linewidth}
            \includegraphics[width=\linewidth]{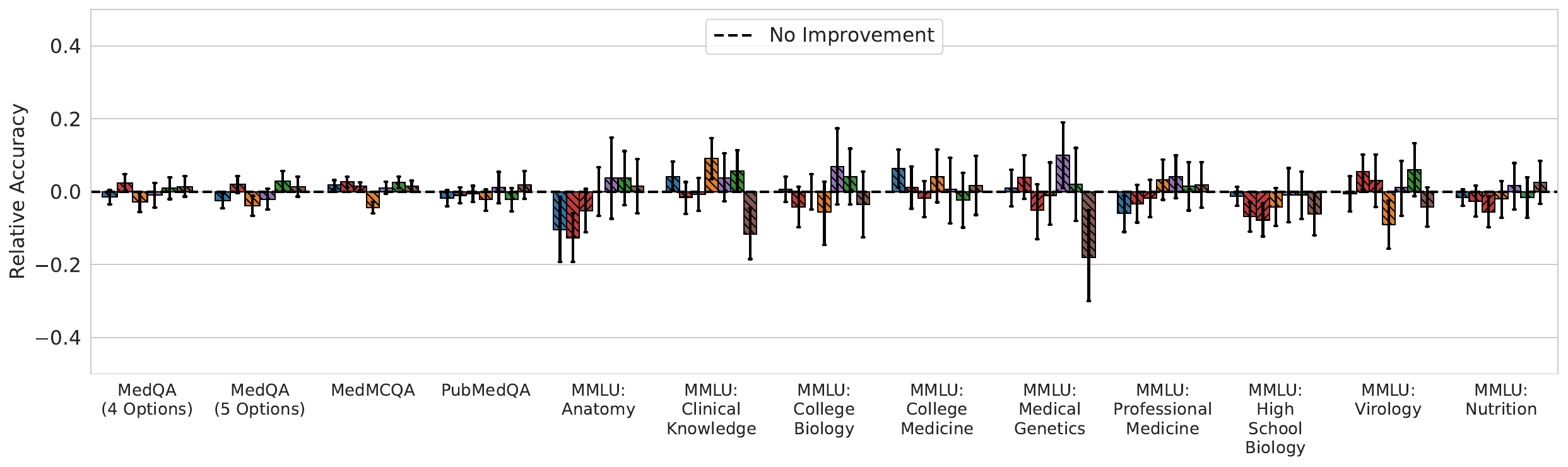}
        \end{subfigure}
    \end{tabular}
    \caption{Medical LLMs do not show a statistically significant improvement over their general-domain counterparts in both (a) zero-shot and (b) 3-shot settings, after independently selecting the best prompt format and examples for each model. Top row shows the absolute exact-match accuracies on the test set, and bottom row shows the relative exact-match accuracies along with 95\% confidence intervals derived via bootstrapping on the test set (see Section~\ref{sec:eval-setup}). We show the results for when model predictions are generated via constrained decoding.}
    \label{fig:llm-logprob-acc-ci}
\end{figure*}

\begin{figure*}[t!]
    \centering
    \begin{tabular}{@{}c@{}c@{\hskip 2pt}c@{}c@{}}
        \multicolumn{4}{c}{
            \begin{subfigure}{0.7\linewidth}
                \includegraphics[width=\linewidth]{figs/vlm-acc-ci-legend-v2.pdf}
            \end{subfigure}
        }
        \\
        \begin{subfigure}{0.03\linewidth}
            \makebox[\linewidth]{\raisebox{60pt}{{(a)}}}
        \end{subfigure} &
        \begin{subfigure}{0.47\linewidth}
            \includegraphics[width=\linewidth]{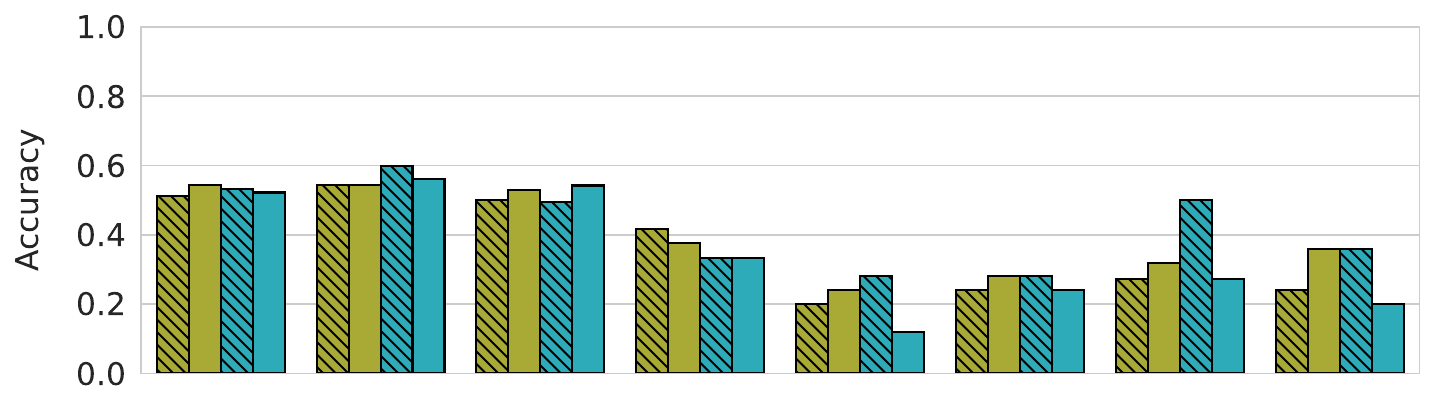}
        \end{subfigure} &
        \begin{subfigure}{0.03\linewidth}
            \makebox[\linewidth]{\raisebox{60pt}{{(b)}}}
        \end{subfigure} &
        \begin{subfigure}{0.47\linewidth}
            \includegraphics[width=\linewidth]{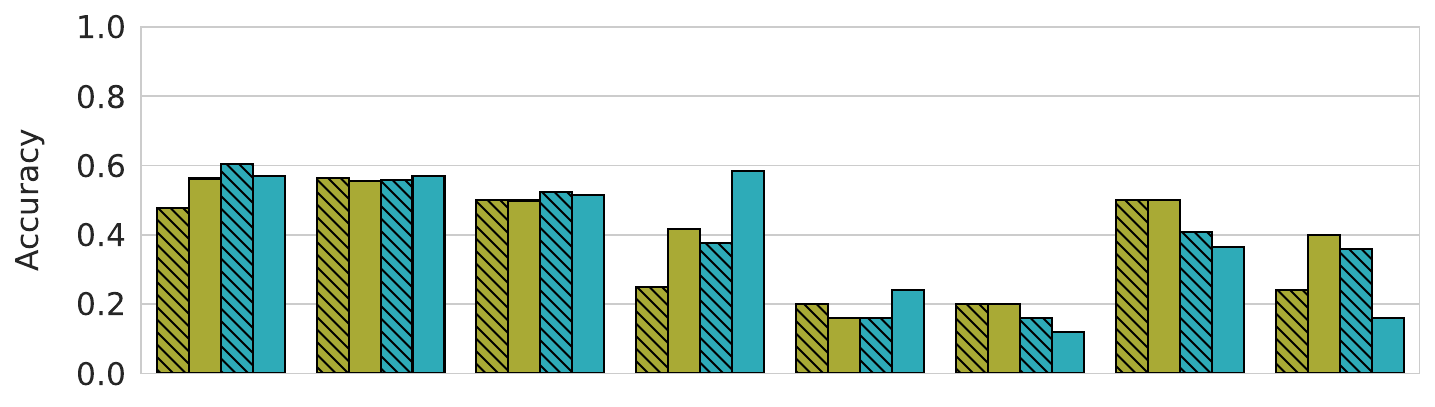}
        \end{subfigure}
        \\
        \begin{subfigure}{0.03\linewidth}
        \end{subfigure} &
        \begin{subfigure}{0.47\linewidth}
            \includegraphics[width=\linewidth]{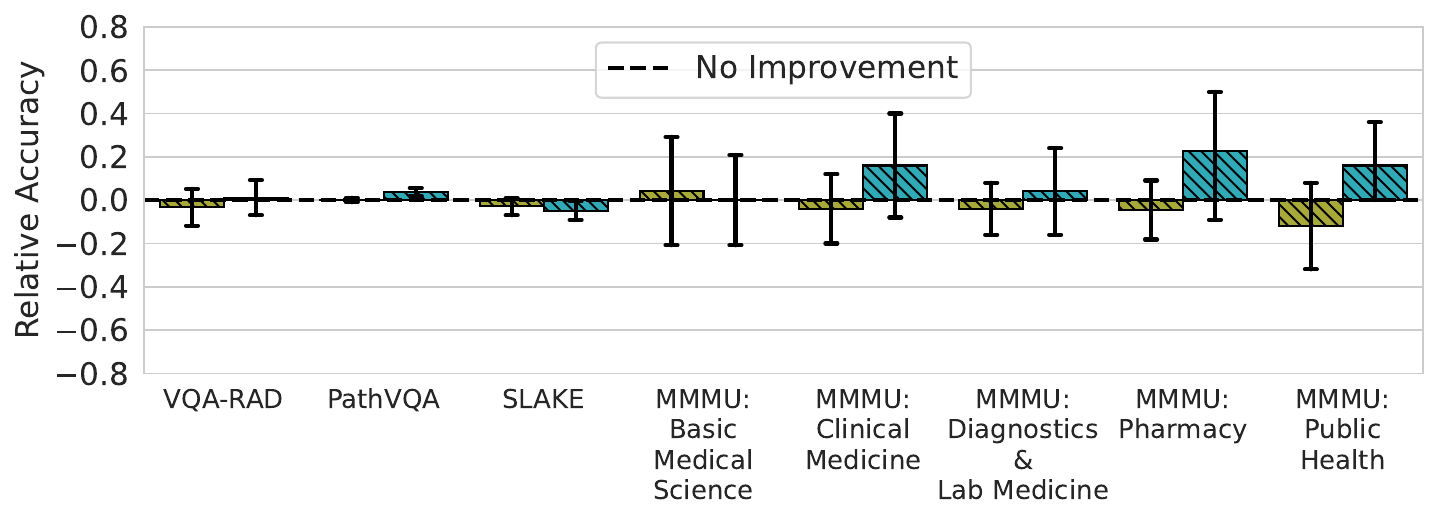}
        \end{subfigure} &
        \begin{subfigure}{0.03\linewidth}
        \end{subfigure} &
        \begin{subfigure}{0.47\linewidth}
            \includegraphics[width=\linewidth]{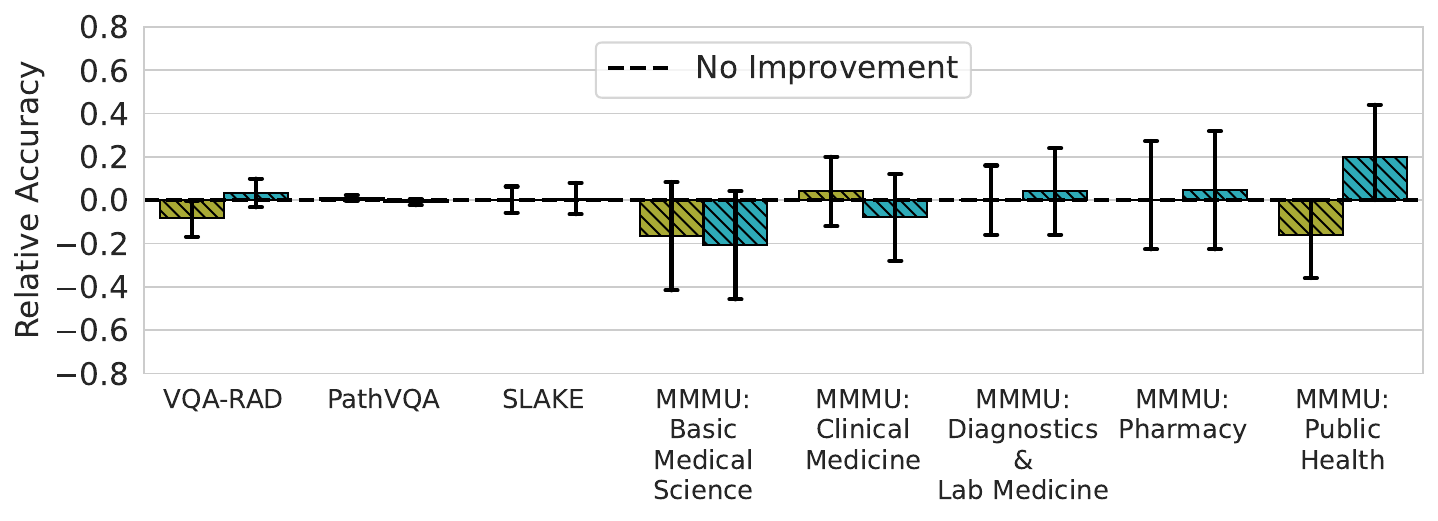}
        \end{subfigure}
    \end{tabular}
    \caption{
    Medical VLMs do not show a statistically significant improvement over their general-domain counterparts in the (a) zero-shot and (b) 3-shot settings, after independently selecting the best prompt format and examples for each model. Top row shows the absolute exact-match accuracies on the test set, and bottom row shows the relative exact-match accuracies along with 95\% confidence intervals derived via bootstrapping on the test set (see Section~\ref{sec:eval-setup}). Here, we show the results for when model predictions are generated via constrained decoding.
    }
    \label{fig:vlm-logprob-acc-ci}
\end{figure*}

Here, we show the constrained decoding results for the medical LLMs and VLMs after independently optimizing the prompt for each model.
In Figures~\ref{fig:llm-logprob-acc-ci}--\ref{fig:vlm-logprob-acc-ci}, we show the absolute and relative exact-match accuracies achieved by all LLMs and VLMs in the (a) zero-shot and (b) 3-shot prompting regimes. 
In Tables~\ref{tab:win-tie-loss-rates-0-logprob}--\ref{tab:win-tie-loss-rates-3-logprob}, we show the zero-shot and 3-shot win/tie/loss rates achieved by each model. 
For \textsc{Clinical-Camel-70B}, we compute the win/tie/loss rates while excluding the MedQA datasets, as discussed in Section~\ref{sec:results}.  
For each medical model, we boldface the win rate if it wins more than it loses to its general-domain base model, and vice versa. 

Figure~\ref{fig:llm-logprob-acc-ci}(a) and Table~\ref{tab:win-tie-loss-rates-0-logprob} show that 4 out of 7 medical LLMs show improvements over their general-domain counterparts in the zero-shot setting, albeit by a limited margin in absolute terms. 
In the 3-shot setting, Figure~\ref{fig:llm-logprob-acc-ci}(b) and Table~\ref{tab:win-tie-loss-rates-3-logprob} show that only 2 out of 7 medical LLMs---\textsc{MediTron-7B} and \textsc{BioMistral-7B}---show improvements over their general-domain counterpart, but with a tie on 92.3\% of all datasets.
For all other models, the win rates are less than or equal to the loss rates, and the majority of cases result in a tie.
Meanwhile, Figure~\ref{fig:vlm-logprob-acc-ci} and Tables~\ref{tab:win-tie-loss-rates-0-logprob}--\ref{tab:win-tie-loss-rates-3-logprob} show that no medical VLM shows a statistically significant improvement over its general-domain counterpart in either the zero-shot or 3-shot setting.

\begin{table}[t!]
    \centering
    \caption{The win, tie, and loss rates (\%) of all medical LLMs (top) and VLMs (bottom) in the zero-shot setting, after independently optimizing the prompts for both medical and general-domain models. Model predictions are generated via constrained decoding.}
    \label{tab:win-tie-loss-rates-0-logprob}
    \resizebox{\linewidth}{!}{
    
    \begin{tabular}{@{}l@{\hskip 7pt}c@{\hskip 7pt}c@{\hskip 7pt}c@{\hskip 3pt}}
        \toprule
        Model & Win & Tie & Loss \\
        \midrule
        \textsc{OpenBioLLM-70B} \citep{OpenBioLLMs} & 7.7 & 76.9 & \textbf{15.4} \\
        \textsc{MediTron-70B} \citep{meditron} & \textbf{30.8} & 46.2 & 23.1 \\
        \textsc{Clinical-Camel-70B} \citep{clinical-camel} & \textbf{18.2} & 72.7 & 9.1 \\
        \textsc{OpenBioLLM-8B} \citep{OpenBioLLMs} & 0 & 53.8 & \textbf{46.2} \\
        \textsc{MediTron-7B} \citep{meditron} & \textbf{23.1} & 76.9 & 0 \\
        \textsc{BioMistral-7B} \citep{biomistral} & \textbf{30.8} & 69.2 & 0 \\
        \textsc{BioMedGPT-LM-7B} \citep{biomedgpt} & 7.7 & 84.6 & 7.7 \\
        \midrule
        \textsc{LLaVA-Med-7B} \citep{llava-med} & 0 & 100.0 & 0 \\
        \textsc{Med-Flamingo-9B} \citep{med-flamingo} & 12.5 & 75.0 & 12.5 \\
        \bottomrule
    \end{tabular}
    }
\end{table}

\begin{table}[t!]
    \centering
    \caption{The win, tie, and loss rates (\%) of all medical LLMs (top) and VLMs (bottom) in the 3-shot setting, after independently optimizing the prompts for both medical and general-domain models. Model predictions are generated via constrained decoding.}
    \label{tab:win-tie-loss-rates-3-logprob}
    \resizebox{\linewidth}{!}{
    
    \begin{tabular}{@{}l@{\hskip 7pt}c@{\hskip 7pt}c@{\hskip 7pt}c@{\hskip 3pt}}
        \toprule
        Model & Win & Tie & Loss \\
        \midrule
        \textsc{OpenBioLLM-70B} \citep{OpenBioLLMs} & 23.1 & 53.8 & 23.1 \\
        \textsc{MediTron-70B} \citep{meditron} & 15.4 & 69.2 & 15.4 \\
        \textsc{Clinical-Camel-70B} \citep{clinical-camel} & 9.1 & 72.7 & \textbf{18.2} \\
        \textsc{OpenBioLLM-8B} \citep{OpenBioLLMs} & 7.7 & 69.2 & \textbf{23.1} \\
        \textsc{MediTron-7B} \citep{meditron} & \textbf{7.7} & 92.3 & 0 \\
        \textsc{BioMistral-7B} \citep{biomistral} & \textbf{7.7} & 92.3 & 0 \\
        \textsc{BioMedGPT-LM-7B} \citep{biomedgpt} & 7.7 & 69.2 & \textbf{23.1} \\
        \midrule
        \textsc{LLaVA-Med-7B} \citep{llava-med} & 0 & 87.5 & \textbf{12.5} \\
        \textsc{Med-Flamingo-9B} \citep{med-flamingo} & 0 & 100.0 & 0 \\
        \bottomrule
    \end{tabular}
    }
\end{table}

\begin{table}[t!]
    \centering
    \caption{The win, tie, and loss rates (\%) of all medical LLMs (top) and VLMs (bottom) in the zero-shot setting, when using a single, fixed prompt optimized only for the medical model. Model predictions are generated via constrained decoding.}
    \label{tab:win-tie-loss-rates-med-0-logprob}
    \resizebox{\linewidth}{!}{
    
    \begin{tabular}{@{}l@{\hskip 7pt}c@{\hskip 7pt}c@{\hskip 7pt}c@{\hskip 3pt}}
        \toprule
        Model & Win & Tie & Loss \\
        \midrule
        \textsc{OpenBioLLM-70B} \citep{OpenBioLLMs} & \textbf{30.8} & 69.2 & 0 \\
        \textsc{MediTron-70B} \citep{meditron} & \textbf{30.8} & 53.8 & 15.4 \\
        \textsc{Clinical-Camel-70B} \citep{clinical-camel} & \textbf{63.6} & 36.4 & 0 \\
        \textsc{OpenBioLLM-8B} \citep{OpenBioLLMs} & \textbf{46.2} & 46.2 & 7.7 \\
        \textsc{MediTron-7B} \citep{meditron} & 7.7 & 76.9 & \textbf{15.4} \\
        \textsc{BioMistral-7B} \citep{biomistral} & \textbf{23.1} & 76.9 & 0 \\
        \textsc{BioMedGPT-LM-7B} \citep{biomedgpt} & \textbf{30.8} & 69.2 & 0 \\
        \midrule
        \textsc{LLaVA-Med-7B} \citep{llava-med} & 0 & 100.0 & 0 \\
        \textsc{Med-Flamingo-9B} \citep{med-flamingo} & \textbf{25.0} & 75.0 & 0 \\
        \bottomrule
    \end{tabular}
    }
\end{table}

\begin{table}[t!]
    \centering
    \caption{The win, tie, and loss rates (\%) of all medical LLMs (top) and VLMs (bottom) in the 3-shot setting, when using a single, fixed prompt optimized only for the medical model. Model predictions are generated via constrained decoding.}
    \label{tab:win-tie-loss-rates-med-3-logprob}
    \resizebox{\linewidth}{!}{
    
    \begin{tabular}{@{}l@{\hskip 7pt}c@{\hskip 7pt}c@{\hskip 7pt}c@{\hskip 3pt}}
        \toprule
        Model & Win & Tie & Loss \\
        \midrule
        \textsc{OpenBioLLM-70B} \citep{OpenBioLLMs} & \textbf{23.1} & 61.5 & 15.4 \\
        \textsc{MediTron-70B} \citep{meditron} & \textbf{7.7} & 92.3 & 0 \\
        \textsc{Clinical-Camel-70B} \citep{clinical-camel} & \textbf{36.4} & 63.6 & 0 \\
        \textsc{OpenBioLLM-8B} \citep{OpenBioLLMs} & \textbf{30.8} & 61.5 & 7.7 \\
        \textsc{MediTron-7B} \citep{meditron} & \textbf{7.7} & 92.3 & 0 \\
        \textsc{BioMistral-7B} \citep{biomistral} & \textbf{7.7} & 92.3 & 0 \\
        \textsc{BioMedGPT-LM-7B} \citep{biomedgpt} & \textbf{30.8} & 69.2 & 0 \\
        \midrule
        \textsc{LLaVA-Med-7B} \citep{llava-med} & \textbf{25.0} & 75.0 & 0 \\
        \textsc{Med-Flamingo-9B} \citep{med-flamingo} & 0 & 100.0 & 0 \\
        \bottomrule
    \end{tabular}
    }
\end{table}

\subsection{Finding 2 (Section~\ref{sec:results})}
\label{sec:constrained-decoding-finding2}

\begin{figure*}[t!]
    \centering
    \begin{tabular}{@{}c@{}c@{}}
        \multicolumn{2}{c}{
            \begin{subfigure}{0.98\linewidth}
                \includegraphics[width=\linewidth]{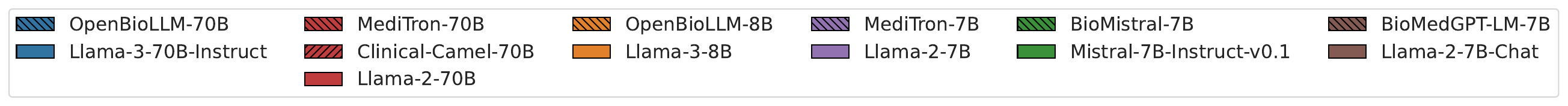}
            \end{subfigure}
        }
        \\
        \begin{subfigure}{0.02\linewidth}
            \makebox[\linewidth]{\raisebox{0pt}{{(a)}}}
        \end{subfigure} &
        \begin{subfigure}{0.96\linewidth}
            \includegraphics[width=\linewidth]{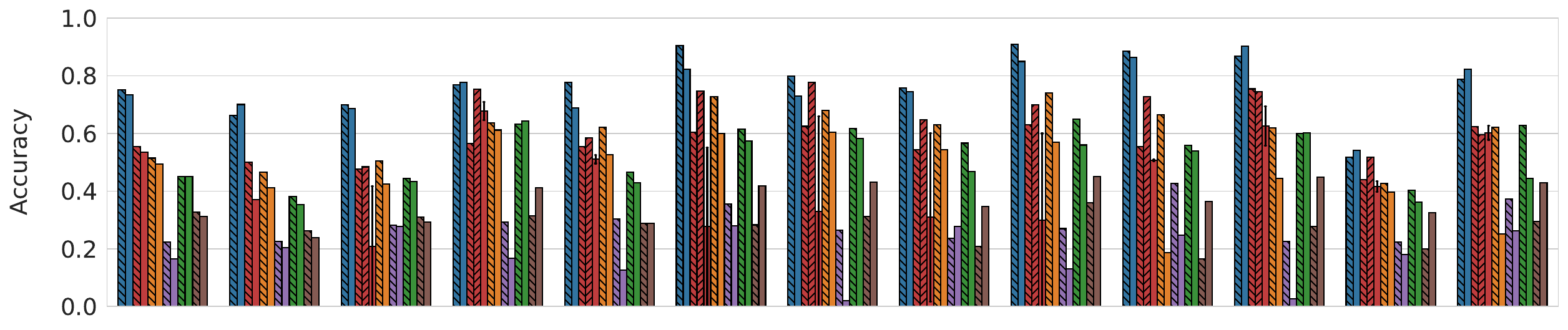}
        \end{subfigure}
        \\
        \begin{subfigure}{0.02\linewidth}
        \end{subfigure} &
        \begin{subfigure}{0.96\linewidth}
            \includegraphics[width=\linewidth]{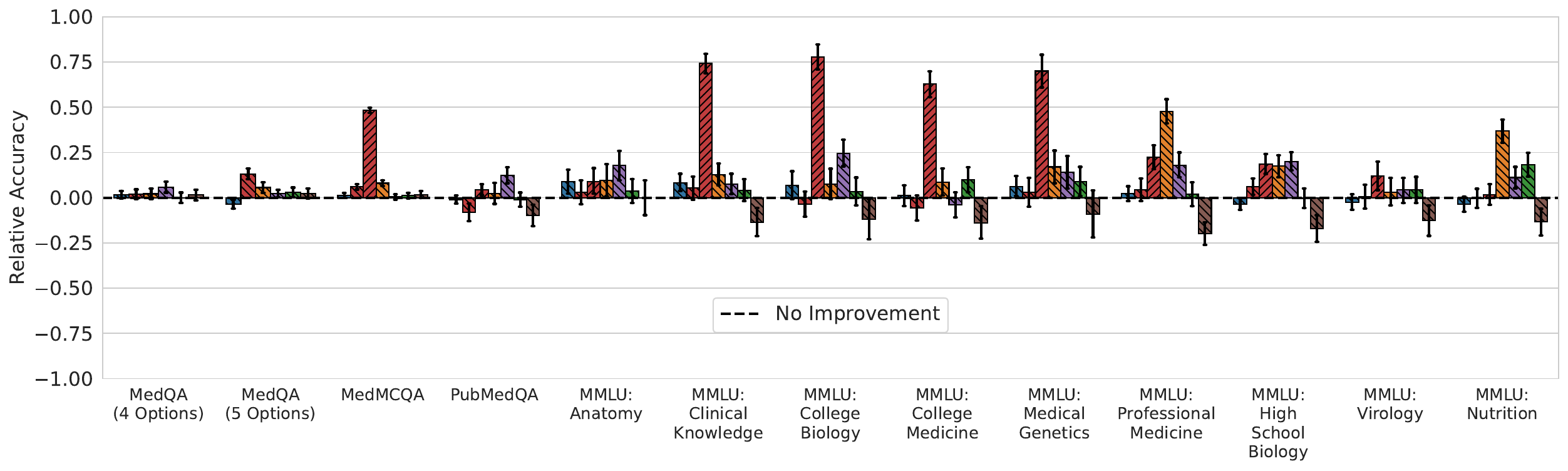}
        \end{subfigure}
        \\
        \begin{subfigure}{0.02\linewidth}
            \makebox[\linewidth]{\raisebox{0pt}{{(b)}}}
        \end{subfigure} &
        \begin{subfigure}{0.96\linewidth}
            \includegraphics[width=\linewidth]{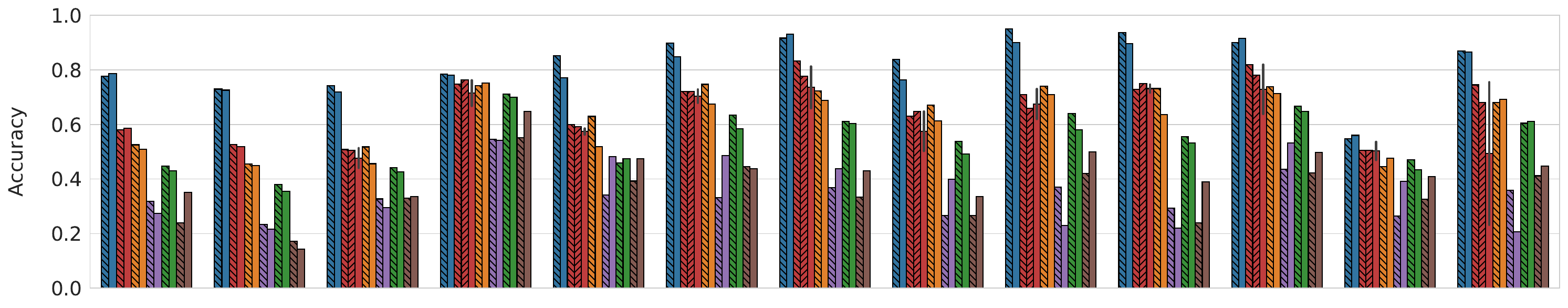}
        \end{subfigure}
        \\
        \begin{subfigure}{0.02\linewidth}
        \end{subfigure} &
        \begin{subfigure}{0.96\linewidth}
            \includegraphics[width=\linewidth]{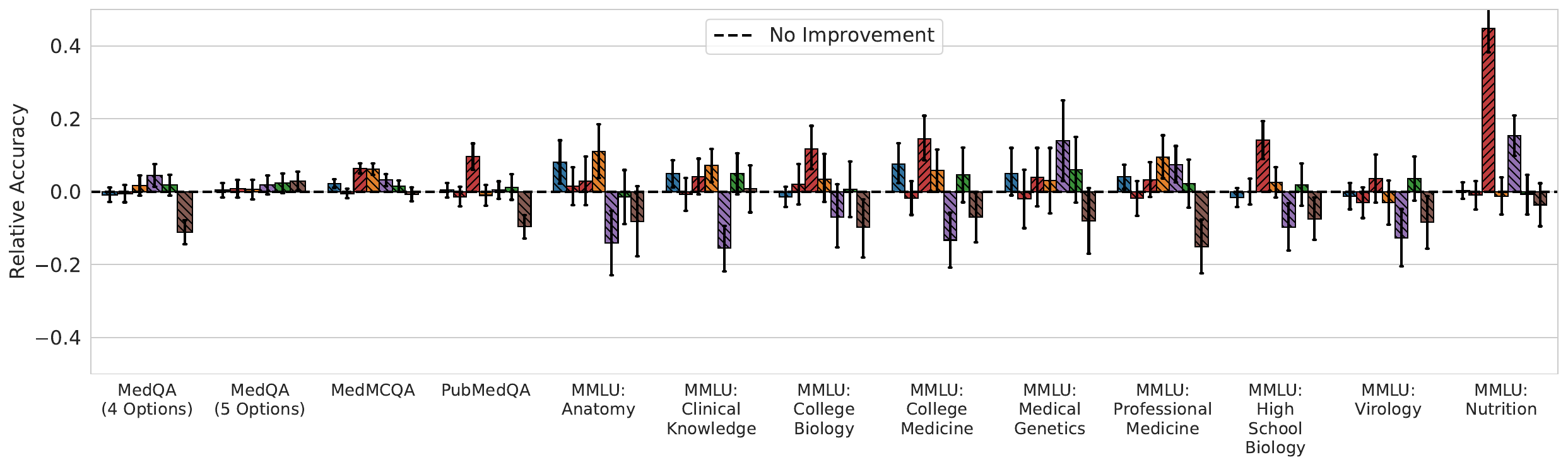}
        \end{subfigure}
    \end{tabular}
    \caption{Using a single, fixed prompt format only optimized for the medical model can overestimate the performance improvements from medical DAPT, in both (a) zero-shot and (b) 3-shot settings. For each setting, top row shows the absolute exact-match accuracies on the test set, and bottom row shows the relative exact-match accuracies along with 95\% confidence intervals derived via bootstrapping on the test set (see Section~\ref{sec:eval-setup}). For \textsc{Llama-2-70B}, which has multiple corresponding medical LLMs (\textsc{MediTron-70B} and \textsc{Clinical-Camel-70B}), we include a min-max error bar in the absolute accuracy plots to show how the absolute accuracy changes with respect to each prompt.
    We show the results for when model predictions are generated via constrained decoding.}
    \label{fig:llm-acc-ci-med-logprob}
\end{figure*}

\begin{figure*}[t!]
    \centering
    \begin{tabular}{@{}c@{}c@{\hskip 2pt}c@{}c@{}}
        \multicolumn{4}{c}{
            \begin{subfigure}{0.7\linewidth}
                \includegraphics[width=\linewidth]{figs/vlm-acc-ci-legend-v2.pdf}
            \end{subfigure}
        }
        \\
        \begin{subfigure}{0.03\linewidth}
            \makebox[\linewidth]{\raisebox{60pt}{{(a)}}}
        \end{subfigure} &
        \begin{subfigure}{0.47\linewidth}
            \includegraphics[width=\linewidth]{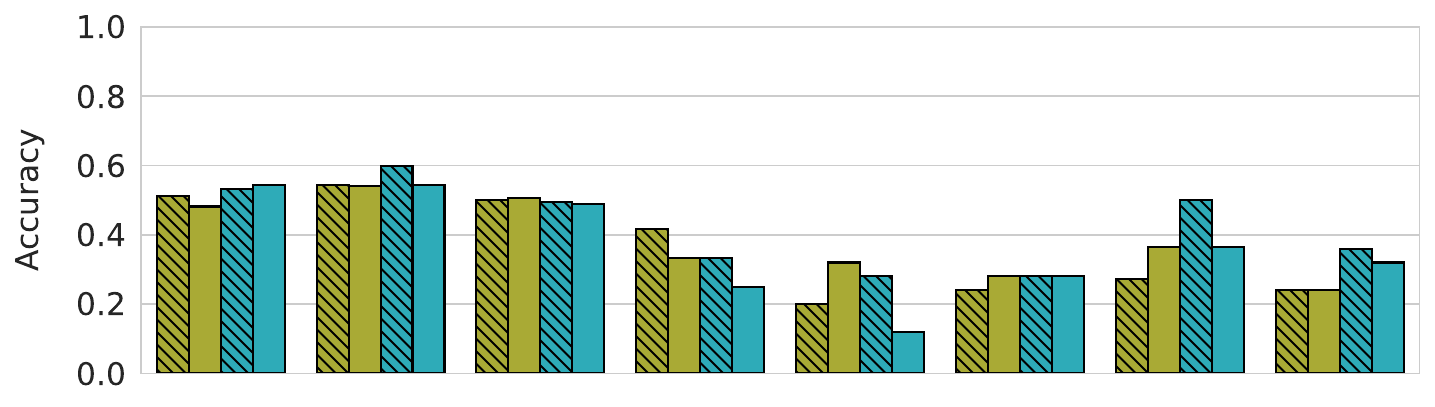}
        \end{subfigure} &
        \begin{subfigure}{0.03\linewidth}
            \makebox[\linewidth]{\raisebox{60pt}{{(b)}}}
        \end{subfigure} &
        \begin{subfigure}{0.47\linewidth}
            \includegraphics[width=\linewidth]{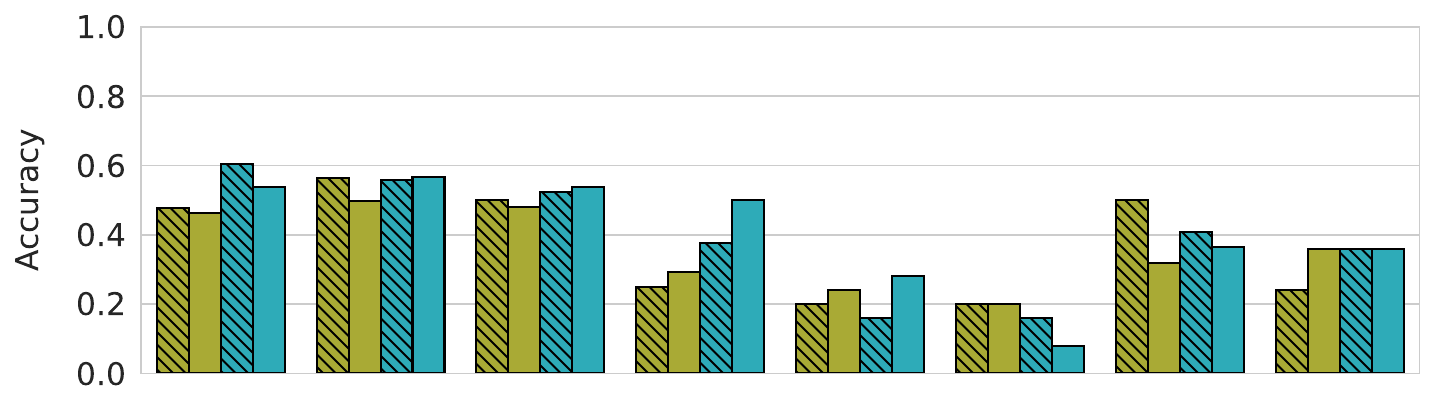}
        \end{subfigure}
        \\
        \begin{subfigure}{0.03\linewidth}
        \end{subfigure} &
        \begin{subfigure}{0.47\linewidth}
            \includegraphics[width=\linewidth]{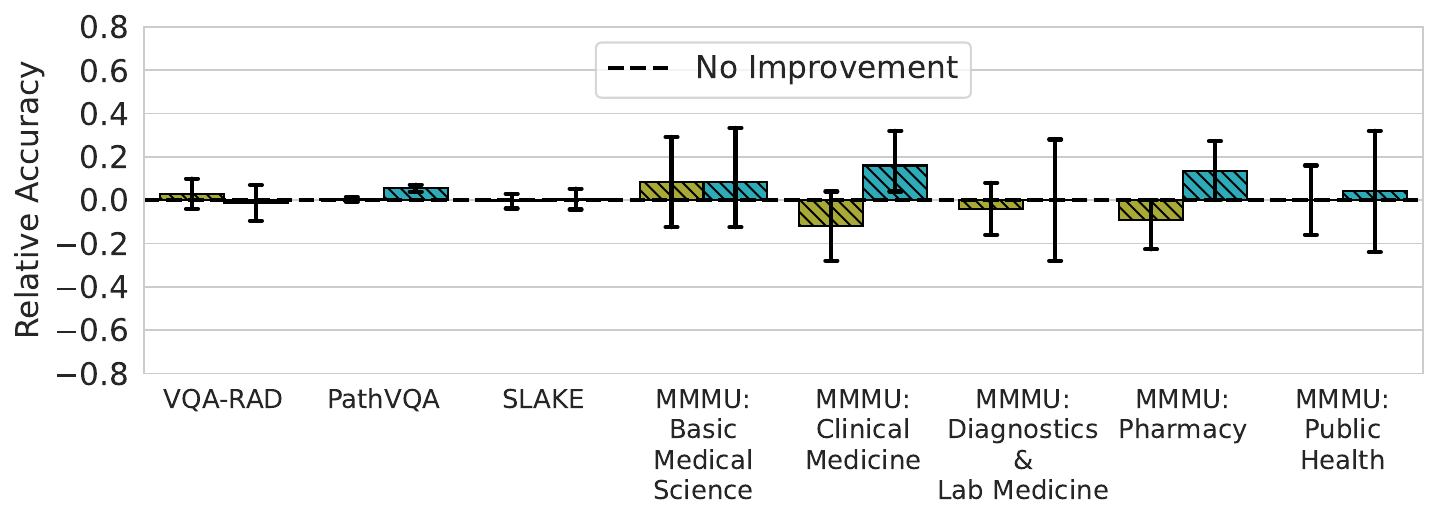}
        \end{subfigure} &
        \begin{subfigure}{0.03\linewidth}
        \end{subfigure} &
        \begin{subfigure}{0.47\linewidth}
            \includegraphics[width=\linewidth]{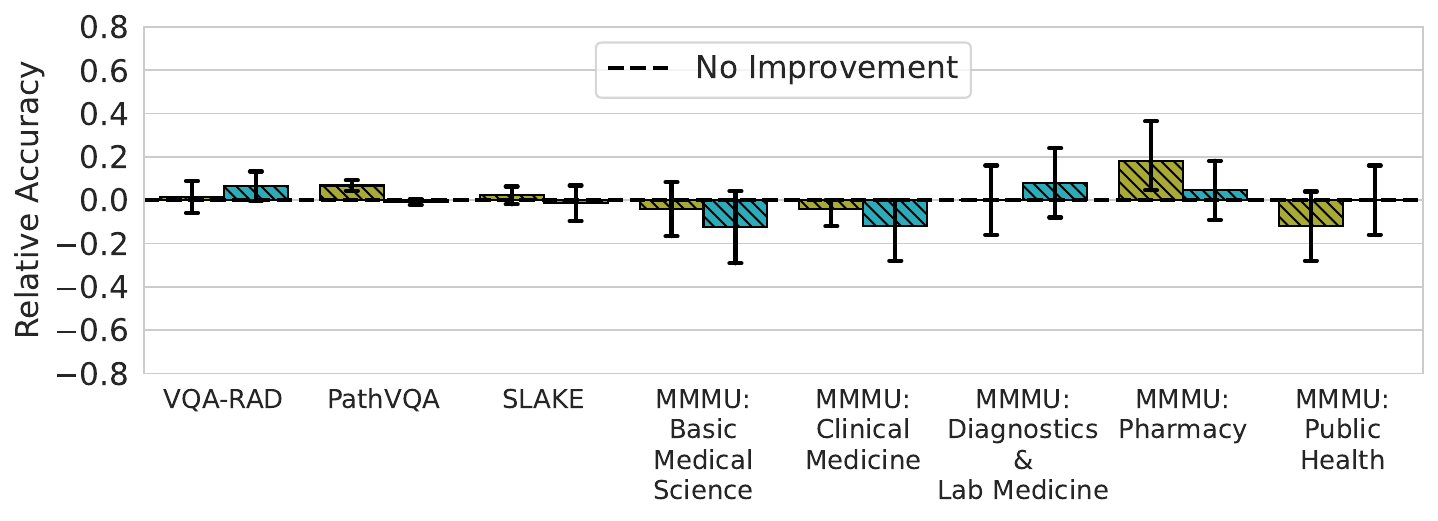}
        \end{subfigure}
    \end{tabular}
    \caption{Using a single, fixed prompt format only optimized for the medical model can overestimate the performance improvements from medical DAPT, in both (a) zero-shot and (b) 3-shot settings. For each setting, top row shows the raw exact-match accuracies on the test set, and the bottom row shows the \textit{relative} exact-match accuracies along with 95\% confidence intervals derived via boostrapping on the test set (see Section~\ref{sec:eval-setup}). Here, we show the results for when model predictions are generated via constrained decoding.}
    \label{fig:vlm-acc-ci-med-logprob}
\end{figure*}

\begin{figure*}[t!]
    \centering
    \includegraphics[width=\linewidth]{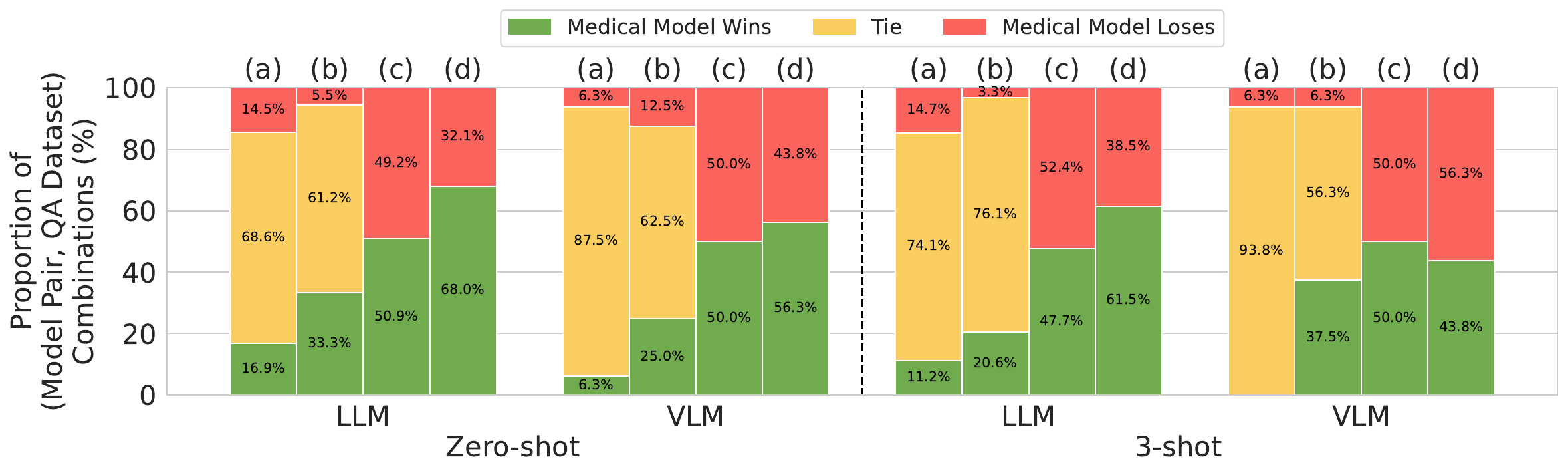}
    \caption{Optimizing the prompt for only the medical model and comparing models without accounting for statistical uncertainty can overestimate the performance improvements from medical DAPT. We show the win/tie/loss rate (\%) of medical models vs. their base models across all (model pair, QA dataset) combinations, when (a) independently optimizing the prompt for each model and performing statistical testing, (b) optimizing the prompt only for the medical model and performing statistical testing, (c) independently optimizing the prompt for each model without statistical testing, and (d) optimizing the prompt only for the medical model without statistical testing. Here, we show the results when model predictions are generated via constrained decoding.}
    \label{fig:opt-logprob-ci-acc}
\end{figure*}

Here, we present the constrained decoding results for medical LLMs and VLMs when using a single, fixed prompt format only optimized for the medical model.
In Figures~\ref{fig:llm-acc-ci-med-logprob}--\ref{fig:vlm-acc-ci-med-logprob}, we show how the absolute and relative exact-match accuracies change for all LLMs and VLMs in the zero-shot and 3-shot settings.  
In Tables~\ref{tab:win-tie-loss-rates-med-0-logprob}--\ref{tab:win-tie-loss-rates-med-3-logprob}, we also show the zero-shot and 3-shot win/tie/loss rates in this scenario. 
For each medical model, we boldface the win rate if it wins more than it loses to its general-domain base model, and vice versa. 

Compared to when the prompt is independently optimized for each model, we see that a greater number of medical models show statistically significant improvements. 
In Figure~\ref{fig:opt-logprob-ci-acc}, we also show how the win/tie/loss rates of the medical models, computed over all (model pair, QA dataset) combinations,
change as we vary the prompting setups as in Finding 2 of Section~\ref{sec:results}. 
As in the greedy decoding setup, we find that for both LLMs and VLMs, the performance improvements from medical DAPT can be substantially overestimated when (i) the prompt is only tailored to the medical model; and (ii) the models are compared only based on their absolute accuracies. 
For example, in the zero-shot setting, the win rate increases from 16.9\% to 68.0\% for medical LLMs and from 6.3\% to 56.3\% for medical VLMs, when only performing prompt selection for the medical model and comparing based on raw absolute accuracy.